\documentclass[10pt,twocolumn,letterpaper]{article}

\PassOptionsToPackage{dvipsnames,table}{xcolor}
\usepackage{algorithm}
\usepackage{algpseudocode} %

\usepackage{graphics}
\usepackage{bm}
\usepackage{multirow}

\usepackage{wacv}      %
\usepackage[accsupp]{axessibility}  %

\makeatletter
\newcommand{\algrule}[1][.2pt]{\par\vskip.5\baselineskip\hrule height #1\par\vskip.5\baselineskip}
\makeatother

\usepackage[scaled=0.88]{helvet} %

\usepackage{mathtools}

\newcommand{\embed}{\mathsf{embed}}
\newcommand{\detect}{\mathsf{detect}}
\newcommand{\embedBlock}{\mathsf{embed\_block}}
\newcommand{\countG}{\mathsf{count_G}}
\newcommand{\detectBlock}{\mathsf{detect\_block}}
\newcommand{\detectChannel}{\mathsf{detect\_channel}}

\newcommand{\roundVal}{\mathsf{round\_val}}
\newcommand{\getSeed}{\mathsf{get\_seed}}

\newcommand{\eval}{\mathsf{evaluate}}
\newcommand{\sk}{\mathsf{sk}} %
\newcommand{\seed}{\mathsf{seed}} %
\newcommand{\score}{\mathsf{score}}
 
\newcommand{\melb}{\textsf{MELB}} %
\newcommand{\perturb}{\mathsf{perturb}}

\usepackage{amsthm}
\theoremstyle{remark}
\newtheorem{remark}{Remark}
\theoremstyle{definition}
\newtheorem{definition}{Definition}
\newtheorem{example}{Example}

\definecolor{wacvblue}{rgb}{0.21,0.49,0.74}
\usepackage[pagebackref,breaklinks,colorlinks,allcolors=wacvblue]{hyperref}

\def\wacvPaperID{413} %
\def\confName{WACV}
\def\confYear{2026}

\title{Where is the Watermark? Interpretable Watermark Detection at the Block Level}

\author{Maria Bulychev, Neil G. Marchant, Benjamin I. P. Rubinstein \\
University of Melbourne\\
{\tt\small \{m.bulychev, nmarchant, benjamin.rubinstein\}@unimelb.edu.au}
}

\begin{document}
\maketitle
\begin{abstract}
Recent advances in generative AI have enabled the creation of highly realistic digital content, raising concerns around authenticity, ownership, and misuse. While watermarking has become an increasingly important mechanism to trace and protect digital media, most existing image watermarking schemes operate as black boxes, producing global detection scores without offering any insight into how or where the watermark is present. This lack of transparency impacts user trust and makes it difficult to interpret the impact of tampering. In this paper, we present a post-hoc image watermarking method that combines localised embedding with region-level interpretability. Our approach embeds watermark signals in the discrete wavelet transform domain using a statistical block-wise strategy. This allows us to generate detection maps that reveal which regions of an image are likely watermarked or altered. We show that our method achieves strong robustness against common image transformations while remaining sensitive to semantic manipulations. At the same time, the watermark remains highly imperceptible. Compared to prior post-hoc methods, our approach offers more interpretable detection while retaining competitive robustness. For example, our watermarks are robust to cropping up to half the image.
\end{abstract}

\section{Introduction}
\label{sec:intro}

Artificial intelligence can generate realistic and high-quality digital content, making it increasingly challenging to distinguish between human-created and machine-generated content. Modern diffusion models like Midjourney, Stable Diffusion, and DALL-E have revolutionised content creation with their ability to create an almost limitless range of novel images, producing everything from photo-realistic imagery to artwork of any style and aesthetic~\cite{glaze}. The spread of AI-generated content poses significant risks, including the potential for misinformation, intellectual property theft, and the blurring of lines between authentic and fabricated media~\cite{risks}. 
For example, the European Union’s AI Act imposes a transparency obligation for generative AI: providers must ensure that AI-generated text, images, video or audio are marked in a machine-readable way and identifiable as artificially generated~\cite{EU_Act}.

Numerous security mechanisms have been developed to provide a means to assert ownership, verify authenticity, and protect intellectual property rights~\cite{buch, steganography_review}. However, most users share personal photos on social media without protection, unaware of the risks this poses. Unprotected images can be easily downloaded and manipulated, allowing malicious actors to create misleading or defamatory content and deepfakes. %
To address these issues, the research community has been exploring various solutions, with watermarking emerging as a prominent technique.

Watermarking involves embedding a unique signal into the original source, which, while subtle enough to remain invisible in normal use, can be extracted to validate multimedia content~\cite{buch}. 
The process consists of two main phases: embedding and detection. In the embedding phase the watermark information is incorporated into the host data using techniques like modifying pixel values~\cite{lsb_wm}, frequency coefficients~\cite{cox}, or other signal properties~\cite{audio_other_signals}. Detection involves analysing the potentially watermarked content to verify the presence of the watermark and extract any hidden information, which can be done through statistical analysis~\cite{kirchenbauer2024watermarklargelanguagemodels}, correlation~\cite{cox}, or pattern-matching methods~\cite{audio_wm}. %

Most image watermarking schemes are treated as black-box systems that output either a binary decision (whether an image is watermarked or not) or a score for the entire input without any explanation~\cite{goals_survey}. This lack of interpretability can be problematic for users who may wish to understand why a particular input is classified as non-watermarked. For instance, an image that was originally generated with a watermark may no longer be classified as such after certain regions are edited. However, existing schemes do not indicate whether the absence of a watermark is due to localised modifications of a watermarked image or due to the whole image not being watermarked in the first place. Yet, many modern watermarking schemes \cite{rivagan, stegastamp, zodiac, tree_rings} rely on deep neural networks (DNNs), which behave like black boxes and offer no clear reasoning for their decisions. Although explainability is widely seen as essential for trustworthy AI~\cite{trustw_ai}, current DNN-based detectors lack reliable methods to provide human-understandable explanations primarily because deep neural networks themselves are not inherently interpretable~\cite{lime_shap}.

This context highlights the need for watermarking schemes that offer region-level interpretability, not only detecting whether tampering has occurred, but also where. 
Targeting this gap, we propose \melb\ Multidimensional Embedding via Localised Blocking---a watermarking method that combines localised embedding, interpretability, and robustness to common image manipulations. Our main contributions are as follows:
\begin{itemize}[leftmargin=*]
    \item We introduce a block-wise image watermarking method that embeds watermark signals into the discrete wavelet transform (DWT) domain using a statistical partitioning approach. This design enables robustness against common manipulations such as compression, noise, and blur, while remaining sensitive to semantically meaningful changes in the image structure (\eg, warping or shape modification).
    \item In contrast to traditional schemes that produce only a global score, our block-wise design allows for the generation of a detection map that highlights the distribution of watermark presence across the image. This enables users to identify which parts of an image have likely been altered and which remain authentic, providing transparency and localised tamper assessment.
    \item Our watermarks are highly imperceptible, with consistently strong results across standard similarity metrics, mostly exceeding prior approaches. Additionally, our embedding process supports explicit control over the trade-off between invisibility and robustness.
    \item Our method demonstrates robustness to common image modifications comparable to state-of-the-art post-hoc watermarking methods and shows notably stronger performance under cropping attacks. Even when up to 50\% of the image is removed, detection remains reliable, with observed true positive rates of up to 91\%.
\end{itemize}

\section{Related Work}
\label{sec:relwork}

\paragraph{Digital Image Watermarking.} 
Traditional watermarking methods \cite{cox, rivagan} embed watermarks post-hoc, working with existing images after their creation. With the development of diffusion models \cite{diffusion_models}, research has expanded to include in-processing watermarking methods \cite{tree_rings, stablesignature} that embed the watermark during the generation process. 
These integrative approaches show promise in embedding signals deep within the image's semantic structure. However, their model-specific nature limits generalisability and applicability to existing images. Our post-hoc approach offers greater flexibility, accommodating AI-generated images from any model as well as natural images.

Post-hoc watermarking techniques \cite{rivagan, SSL, CIN, stegastamp, zodiac} can be further distinguished between spatial domain and transform domain methods \cite{survey_2008, review_2020}. Additionally, deep learning-based methods extend beyond these domains by leveraging latent feature spaces for watermark embedding~\cite{stablesignature, jigmark}.
Spatial domain methods \cite{zhang_spatial} operate directly on pixel values, offering low computational complexity but generally lower robustness to image processing operations~\cite{buch, review_2020}.
The spatial domain refers to the representation of an image as an array of pixels %
\cite{zhang_spatial}. 
By contrast, transform domain methods, such as those using the Discrete Wavelet Transform (DWT)~\cite{Xia:98_dwt}, typically exhibit higher robustness to image alterations and compression.

Recent advances in deep neural networks (DNNs) have led to learning-based watermarks that push the boundaries of robustness and imperceptibility. StegaStamp~\cite{stegastamp} embeds invisible messages into photographs by training with differentiable image perturbations for noise robustness and using a spatial transformer to account for geometric distortions from printing and re-capturing. RivaGAN~\cite{rivagan} is a video watermarking method that uses an attention-guided dual-stream architecture to learn optimal watermark placement. The method's robustness is enhanced via adversarial training, where a critic ensures visual quality and a removal network continuously challenges the watermark's integrity. Other recent work ~\cite{iconmark, explanation_wm_shao} explores interpretable and robust watermarking for ownership verification.
While these DNN-based approaches offer strong robustness, they lack the ability to provide localised detection, which is a key feature of our proposed method.

\paragraph{Localised Watermarking.} Despite the potential for enhanced interpretability and forensic utility, localised image watermarking—where a watermark's presence can be verified for specific image regions—has seen limited research to date. An early example is the hierarchical block-based approach of \citeauthor{celik2002}~\cite{celik2002}, which introduced fragile watermarks with block-level tamper localisation using least significant bit embedding. More recently, EditGuard~\cite{editguard} introduced a dual watermarking strategy that embeds both localisation and copyright watermarks. While effective for tamper detection, this method is not robust to geometric augmentations such as cropping. Building on this, in concurrent work \citeauthor{WAM}~\cite{WAM} proposed WAM, a localised watermarking method achieving strong robustness. However, WAM is designed to preserve the watermark even after severe alterations that significantly degrade the image, a feature that can be undesirable for image forensics. Our watermark, in contrast, is intentionally fragile and designed to break under such strong editing, providing a clearer signal that the image has been heavily manipulated. Furthermore, while pre-trained models for WAM are available, its detector is a transformer model with approximately 100 million parameters. Our method is training-free, using a lightweight detector based on projections in DWT space. In contrast to both EditGuard and WAM, our work thus offers a computationally efficient approach to localised detection that is robust to common manipulations but fragile to severe, perceptually obvious ones. The concept of localised detection has seen limited exploration in other domains as well. For instance, AudioSeal \cite{audioseal_interpret_wm} tackles localised watermarking for AI-generated speech, providing sample-level resolution in longer recordings.

\paragraph{Statistical Detection of Randomised Partitions.} \citeauthor{kirchenbauer2024watermarklargelanguagemodels}~\cite{kirchenbauer2024watermarklargelanguagemodels} introduced a simple but innovative technique that embeds a watermark into the output of large language models by subtly modifying token probabilities during generation. This is achieved by partitioning the vocabulary at each step evenly into two distinct groups (“Green List” and “Red List”), where the likelihood of Red List tokens appearing in the generated text is reduced, while Green List tokens are boosted. 
A similar watermark was proposed for tabular data \cite{tabularmark} where the watermark is embedded by perturbing specific cells (``key cells''), selecting the perturbation values from the ``Green Domains''. 
Our work builds upon these statistical partitioning concepts, adapting them to the image domain. This approach enables us to achieve localised detection, a feature notably absent in most existing image watermarking schemes, while maintaining a balance between imperceptibility and robustness.

\begin{figure*}[t]
    \centering
    \includegraphics[height = 6cm]{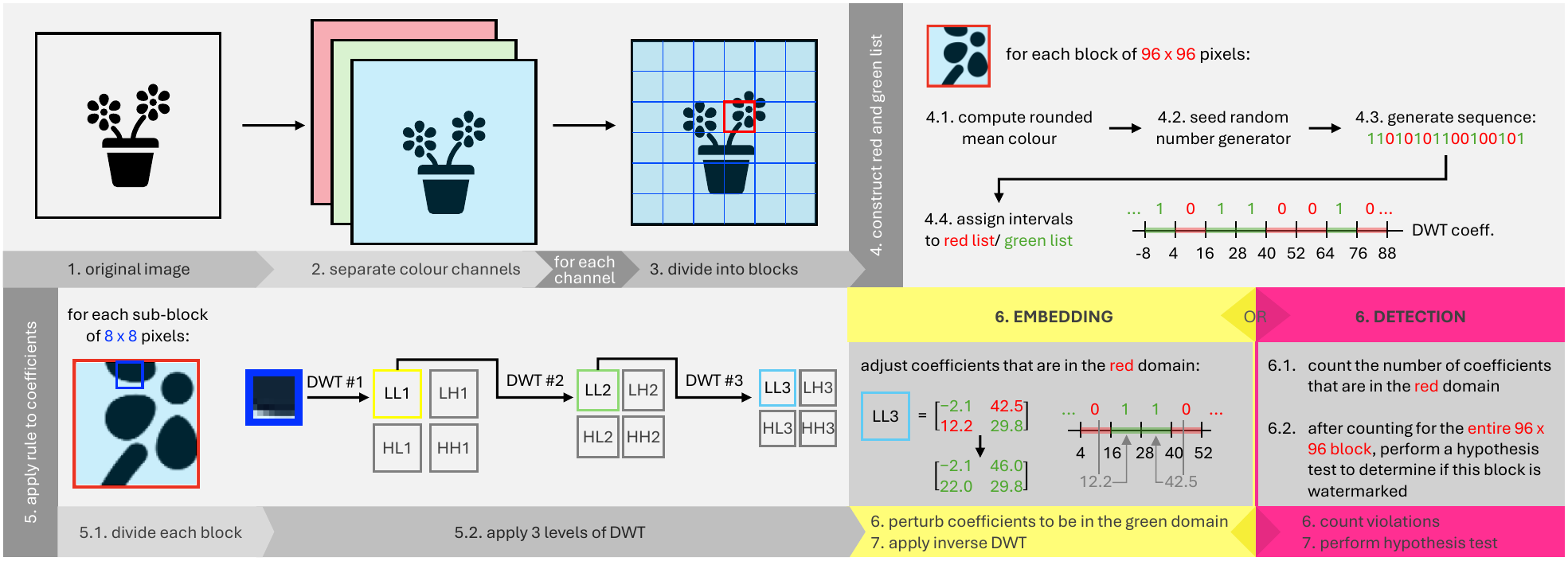} %
    \caption{Pipeline for our watermarking method using $l=12, k=8, m=96, d=3$. The embedding phase partitions the domain of DWT coefficients for each colour channel into allowed and disallowed regions. It then perturbs these coefficients to ensure they all fall within allowed regions. The detection phase reverses this process and counts rule violations. Components highlighted in light grey are identical in both embedding and detection procedures.}
    \label{fig:emb_det_process} %
\end{figure*}

\section{Preliminaries}

Our watermarking framework embeds a robust, yet imperceptible signal into images, allowing owners to identify their intellectual property without compromising visual quality. It also provides an interpretable detection map that visually highlights areas of potential modification (such as through insertions, deletions, or warping).
This addresses scenarios on image-sharing platforms like Facebook or Instagram, where users post content without protection, risking unintended alterations or redistribution \cite{proactive_semi_fragile}.

The framework is designed to maintain watermark integrity under common image editing operations. This robustness ensures that the connection to the original creator is preserved even when the image undergoes benign modifications or is subject to attempts at removal, aiding in authentication and traceability.

Our treatment of digital image watermarking introduces an additional aspect: localised detection. 
Unlike traditional watermarking schemes that provide only a global decision for an entire image, our formulation allows for fine-grained, interpretable watermark detection at the local level.
We focus on post-hoc watermarking, where watermarks are embedded into any image, regardless of its origin---traditional, photographic, or AI-generated---offering broad utility in diverse scenarios.

\subsection{Problem Formulation}

An image is a three-dimensional array $I \in \mathcal{V}^{h \times w \times c}$ where $h$ is the height, $w$ is the width, $c$ is the number of channels, and $\mathcal{V}$ is the range of intensity values. 
For 24-bit colour images, $\mathcal{V} = \{0, 1, \ldots 255\}$ and $c = 3$ (\eg, corresponding to RGB).
We write $I_{i,j,k}$ to refer to the intensity of the $k$-th channel at spatial location $(i, j)$, where $i$ indexes rows (top to bottom) and $j$ indexes columns (left to right). 

An image watermarking scheme consists of two fundamental algorithms: an embedding algorithm and a detection algorithm. 
These algorithms are typically operated by a single entity, which we will call the \emph{watermarker}. 

\begin{definition}
The embedding algorithm maps an original image $I \in \mathcal{V}^{h \times w \times c}$ to a watermarked image $I^\star = \embed(I, \sk)$ using a secret key $\sk$ known only to the watermarker. 
\end{definition}

\begin{definition}
The detection algorithm $\detect(\tilde{I}, \sk)$ takes a possibly watermarked image $\tilde{I} \in \mathcal{V}^{h \times w \times c}$ and secret key $\sk$. 
This yields two outputs: a \emph{global detection} score $d \in [0, 1]$ which indicates whether the entire image is watermarked, and a \emph{localised detection map} $V \in [0, 1]^{h \times w}$ quantifying the likelihood that localised regions are watermarked. 
These scores can be thresholded, so that a positive detection is represented by 1, while a negative detection is represented by 0.
\end{definition}

We now summarise the design desiderata for our image watermarking scheme:
\begin{itemize}
    \item \emph{Imperceptibility}: The watermarked image $I^\star$ should be visually indistinguishable from the original image $I$. 
    Formally, for some perceptual distance function $d$ and threshold $\epsilon$, we require $d(I, I^\star) \leq \epsilon$.
    \item \emph{Blindness}: The watermark should be detectable without access to the original image.
    \item \emph{Localised detection}: Watermark detection should highlight localised factors that influence detection decisions.
    \item \emph{Soundness}: The false positive rate $\delta$ of the detection algorithm should be negligible. For $\mathcal{I}$ a 
    distribution of non-watermarked images, $\Pr_{I \sim \mathcal{I}}[\detect(I, \sk) \neq 0] < \delta$ where we are implicitly applying a threshold to global detection scores.
    \item \emph{Robustness}: The watermark should be detectable even after common image transformations $T$:  
    $\detect(T(I^\star), \sk) \approx \detect(I^\star, \sk)$.
\end{itemize}

\paragraph{Threat Model}
To define the scope of our threat model, we describe the abilities of attackers as follows. 1)~We assume that attackers have full access to watermarked images and 2)~are aware of the general process of the watermark embedding and detection algorithms, but 3)~do not know secret information, such as the secret key (see \cref{sec:embedding}). Moreover, we assume that potential attackers 4)~aim to modify images while preserving their visual quality. For example, when adding noise, the modifications are expected to remain subtle and imperceptible to the human eye. In cases of blurring or compression, the objective is to maintain the image's core content and semantics. An observer should still be able to recognise and understand the primary subject.

\section{\texorpdfstring{\melb}{MELB}: Multidimensional Embedding via Localised Blocking}
\label{sec:algorithms}
In this section, we present the algorithms of our watermarking method in detail. The process is illustrated in
\cref{fig:emb_det_process} and involves two stages: embedding and detection. %

\subsection{Watermark Embedding} \label{sec:embedding}

Our watermark embedding method operates in the transform domain, specifically using the Discrete Wavelet Transform (DWT). The key idea is based on the concept of statistical partitioning of DWT coefficients into allowed (``green") and disallowed (``red") regions. 
Further background on the DWT is provided in App.~\ref{app:dwt}, and pseudocode is provided in Algorithm~\ref{alg:MELB} of App.~\ref{app:algorithms}.

We first divide the image into non-overlapping blocks of size $m \times m$, ignoring incomplete edge blocks. 
For each block, we generate a seed using a generic $\getSeed$ function (\eg, in our experiments, this function returns the rounded mean intensity of the block per channel).
This seed, along with a random secret key $\sk$, is passed as input to a pseudorandom function (PRF), which returns a pseudorandom binary sequence. 
The resulting binary sequence determines the partitioning of DWT coefficients for the block into red~(0) or green~(1). 

The actual embedding occurs at the sub-block level, using $k \times k$ sub-blocks ($k \leq m$) within each larger block. 
This reduces the visibility of block-wise artefacts in the watermarked image, as larger blocks could introduce more noticeable discontinuities at boundaries. 
For each sub-block:
\begin{enumerate}
    \item We map the image to the transform domain by applying $d$ levels of DWT to the LL band.
    \item Using the parent block's red\slash green partitioning, we project the DWT coefficients onto the allowed (green) region by perturbing any coefficient in a red interval to the centre of the nearest green interval. This perturbation is applied only to the LL (low-frequency) band at level $d$, as low-frequency components are expected to be more robust to common image manipulations.
    \item Finally, we apply the inverse DWT $d$ times to obtain the watermarked sub-block.
\end{enumerate}

This process ensures that in the watermarked image, nearly all DWT coefficients fall within green regions, while in natural, non-watermarked images, approximately 50\% of the coefficients would fall in each region. 
App.~\ref{app:parameters} provides a detailed discussion of the parameters, their control mechanisms, and our design choices. %
The key parameters impacting visibility and robustness are the interval length $l$ and the number of DWT levels $d$. Increasing 
$l$ generally improves robustness but may introduce visual artefacts. Higher values of $d$ are typically more advantageous as they enhance both imperceptibility and robustness.
Optionally, our method can operate in a mode that skips watermark embedding in low-entropy blocks, where perturbations might be more perceptible (see App.~\ref{app:embedding_settings}).

\begin{figure*}
    \centering
    \includegraphics[width=\textwidth]{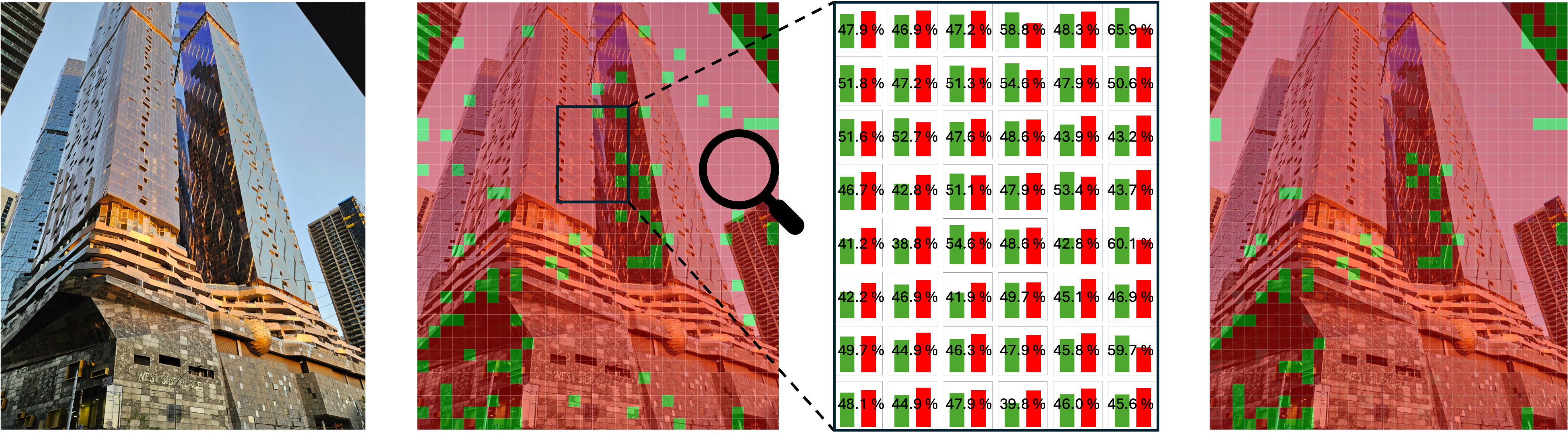} %
    \caption{We apply our detection method to a non-watermarked image (left panel) and show the resulting localised detection map (middle-left panel). The middle-right panel illustrates the red/green coefficient distribution for a small part of the image. Most blocks show a balanced red/green ratio ($\sim$50/50), with minor deviations towards the distribution tails. False positives are effectively filtered out, resulting in only 8.3\% of the area being classified as watermarked.
    The right panel shows the post-processed detection map (see \cref{sec:Map}).} \label{fig:detection_example}
\end{figure*}

Our block-based approach directly supports our goal of localised detection. 
As we shall see in the next section, the detection algorithm outputs scores for each block, which are statistically significant due to the multiple sub-blocks contained within. The resolution of our detection map is controlled by the block size $m$, allowing for fine-grained localisation of potential modifications.

\begin{remark} \label{remark 1}
The use of a PRF with a random secret key $\sk$ mitigates forgery risks by preventing unauthorised creation of watermarked images. 
This approach necessitates a more centralised setup for detection---\eg, where the detection algorithm is made accessible to users through a public API. 
The PRF can be instantiated with a block cipher like AES or a cryptographic hash function such as SHA3 \citep{kirchenbauer2024watermarklargelanguagemodels}. 
While our experiments use a fixed secret key, employing multiple keys can further strengthen the method against brute-force attacks (see \cref{private_multiple_keys}).
\end{remark}

\subsection{Watermark Detection} \label{sec:Detection}

To check for the presence of a watermark in an image, we essentially reverse the embedding process. 
The complete procedure is detailed in Algorithm~\ref{alg:MELB_detection} of App.~\ref{app:algorithms}. 
Here we describe the key steps.
For each colour channel of each block, we:
\begin{enumerate}
    \item Compute the seed using the $\getSeed$ function. 
    This function should produce a stable seed even when the block has undergone minor modification, so that we can recover the seed that was used to embed the potential watermark without access to the original image.
    \item Use the seed, along with the PRF and secret key $\sk$, to determine the red\slash green partitioning of DWT coefficients for the block.
    \item Run detection for each $k \times k$ sub-block: we apply $d$ levels of DWT to the LL sub-band, then examine the LL sub-band at the last level, counting how many coefficients fall in the green (allowed) region.
\end{enumerate}

From this block-level analysis, we can proceed to compute either a localised detection map or a global score for the entire image. 
We describe each of these processes in more detail below.

\paragraph{Localised Detection Map.}
For each $m \times m$ block, we aggregate the count of coefficients in the green region across all its sub-blocks. We set a detection threshold on this aggregate count based on a statistical hypothesis test with a prescribed significance level (see App.~\ref{app:hyp_test} for details). 
The resulting detection map is generated at the block level: each $m \times m$ block is coloured green (indicating it is watermarked) if its aggregate green count exceeds the threshold, otherwise it is coloured red, as shown in \cref{fig:detection_example,fig:inserted_couple}.

\paragraph{Global Detection.}
While the localised detection map provides detailed information about the distribution of the watermark, a global measure is important for making a definitive assessment about whether the entire image is watermarked.
To achieve this, we aggregate over the localised detection map, counting the fraction of blocks classified as watermarked (green).
If the total count exceeds a predetermined threshold (calibrated to achieve a prescribed false positive rate), we predict that the entire image is watermarked; otherwise, we conclude it is not watermarked.

\begin{remark}
Since we embed the watermark into blocks throughout the image, we can implement a detection method that identifies watermarks even in cropped images. 
The method is based on a brute-force strategy to systematically search for the correct starting point of the embedded blocks, as described in \cref{sec:cropping}.
\end{remark}

\subsubsection{Interpreting the Localised Detection Map} \label{sec:Map}

The localised detection map provides valuable insights beyond global detection. 
It can be a helpful aid for image forensics, allowing users to identify specific locations where a watermarked image may have been altered or where protected (watermarked) content has been inserted into a non-watermarked image.

It is important, however, to acknowledge that false positives and false negatives can occur since the hypothesis test suffers some error depending on the chosen significance level. To enhance the visual clarity of our map we apply an optional post-processing step, akin to a low-pass filter: if a green block is entirely surrounded by red blocks, we consider it a likely false positive and recolour it red. Similarly, red blocks completely surrounded by green blocks are recoloured green. It is important to note that this post-processing is only done to improve the utility of the detection map as a visual aid---we do not apply it when computing the global score for the image.

\paragraph{Example: Non-watermarked Image.}
\cref{fig:detection_example} displays a non-watermarked image alongside its localised detection map and an analysis of its DWT coefficients. 
The analysis (zoomed in panel) shows the frequency of rule violations averaged across the three colour channels.
We observe that most blocks (over 91\%) have balanced distributions of red and green domain coefficients. 
This balance is expected in natural non-watermarked images where approximately half of the blocks exhibit more than 50\% violating coefficients, while the other half show less than 50\%.

The raw localised detection map (to the right of the image) visually represents the result of our detection process. We see that most of the image's area is red, with only small, scattered green clusters. These green blocks, which make up only 8.3\% of the total, and are scattered rather randomly across the image, represent natural statistical variations rather than the presence of a watermark.

It is worth noting that small clusters require careful interpretation. Dense clusters may naturally form in highly monochromatic areas with minimal variation, such as the black shadow in the top right corner of our example. In these regions, blocks tend to have similar block seeds (mean colours) and therefore similar DWT coefficients. If these coefficients happen to fall in the ``watermarked'' regions of our detector, a green cluster may appear even in non-watermarked images.

\begin{figure}[t]
    \begin{subfigure}[t]{\linewidth}
        \hfill
        \includegraphics[height=2.7cm]{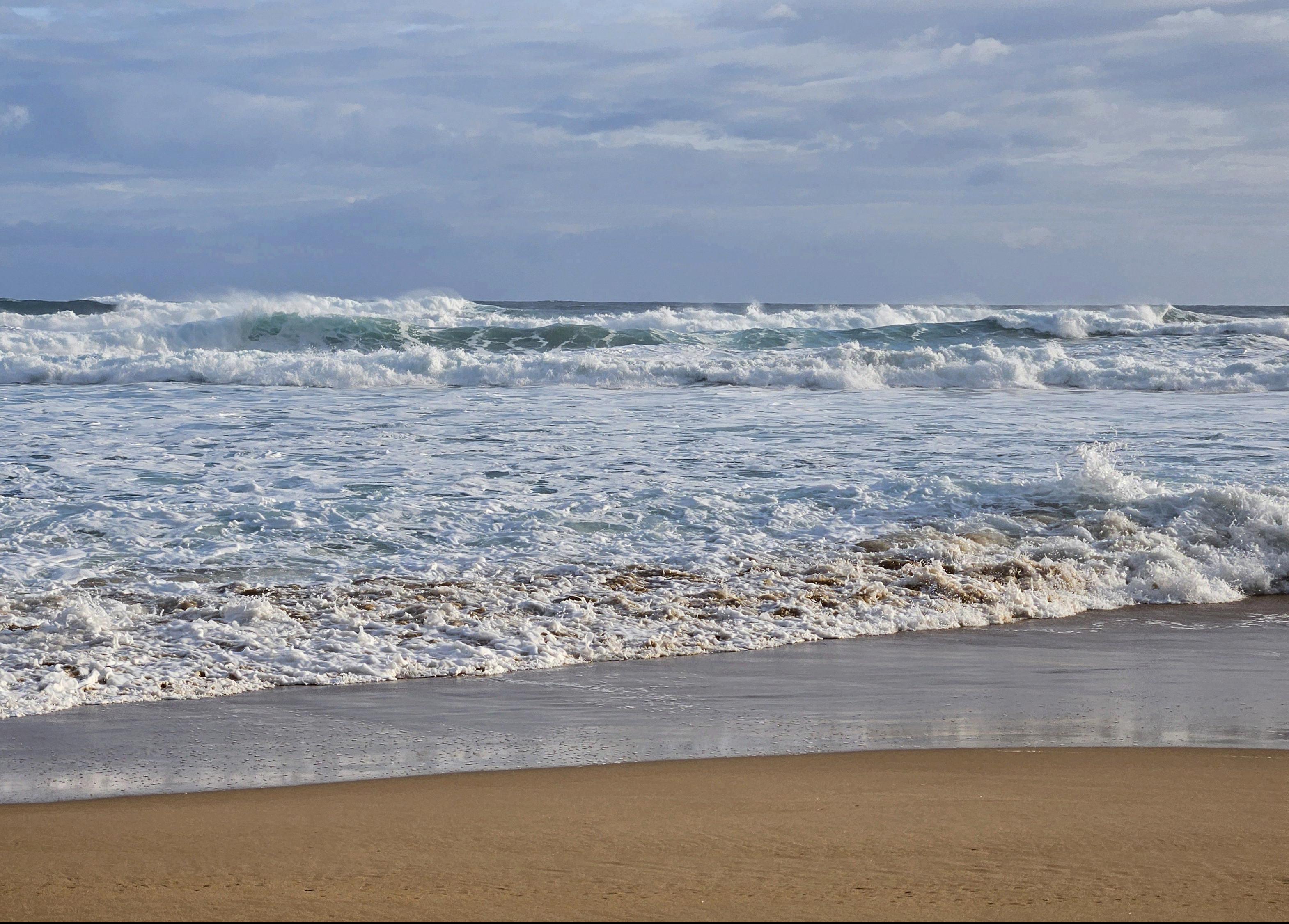}
        \hfill
        \includegraphics[height=2.7cm]{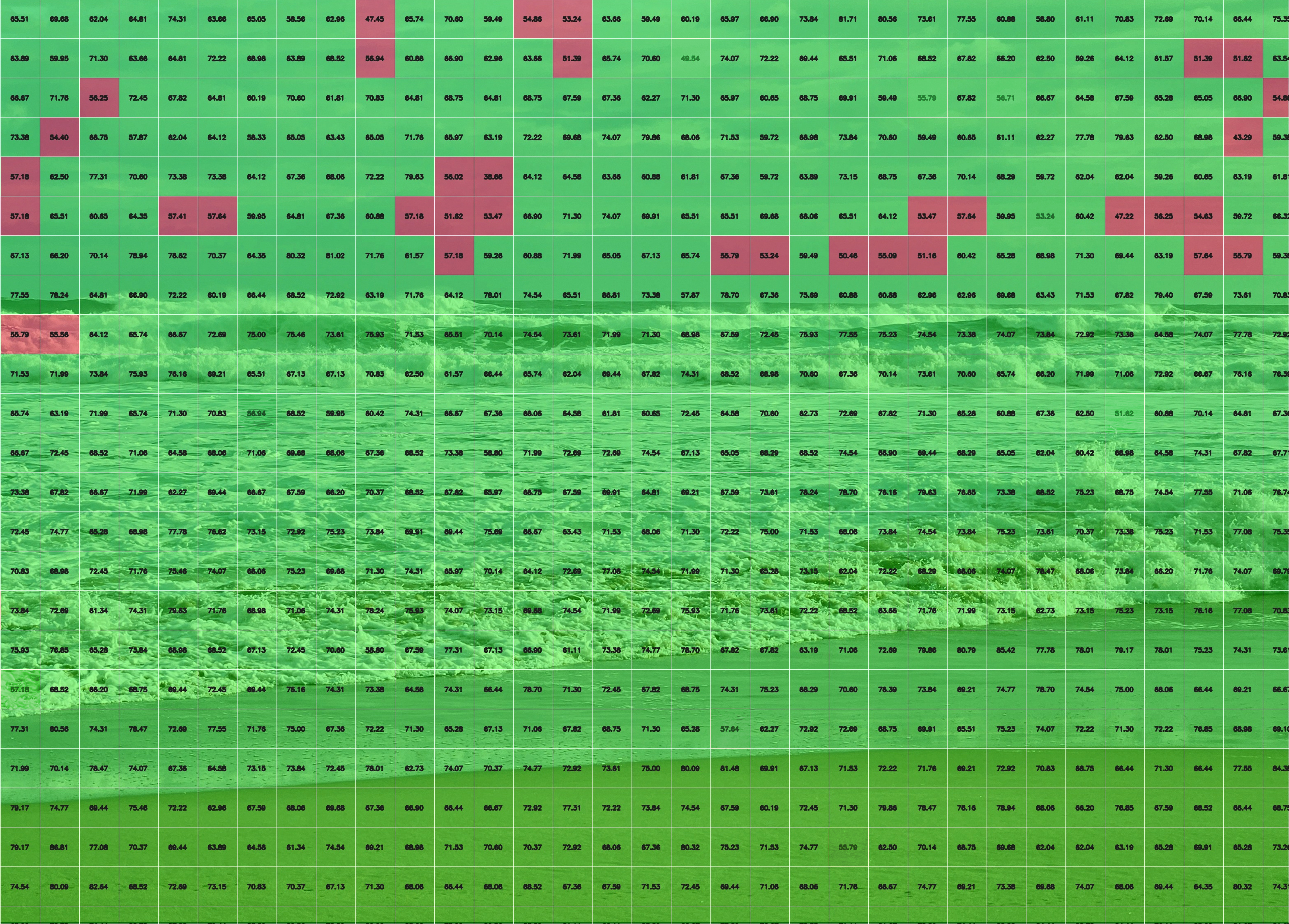}
        \hfill
        \null
        \caption{Watermarked image and the corresponding localised detection map generated during watermark detection process. } \label{watermarked_beach}
    \end{subfigure}
    \begin{subfigure}[t]{\linewidth}
        \hfill
        \includegraphics[height=2.7cm]{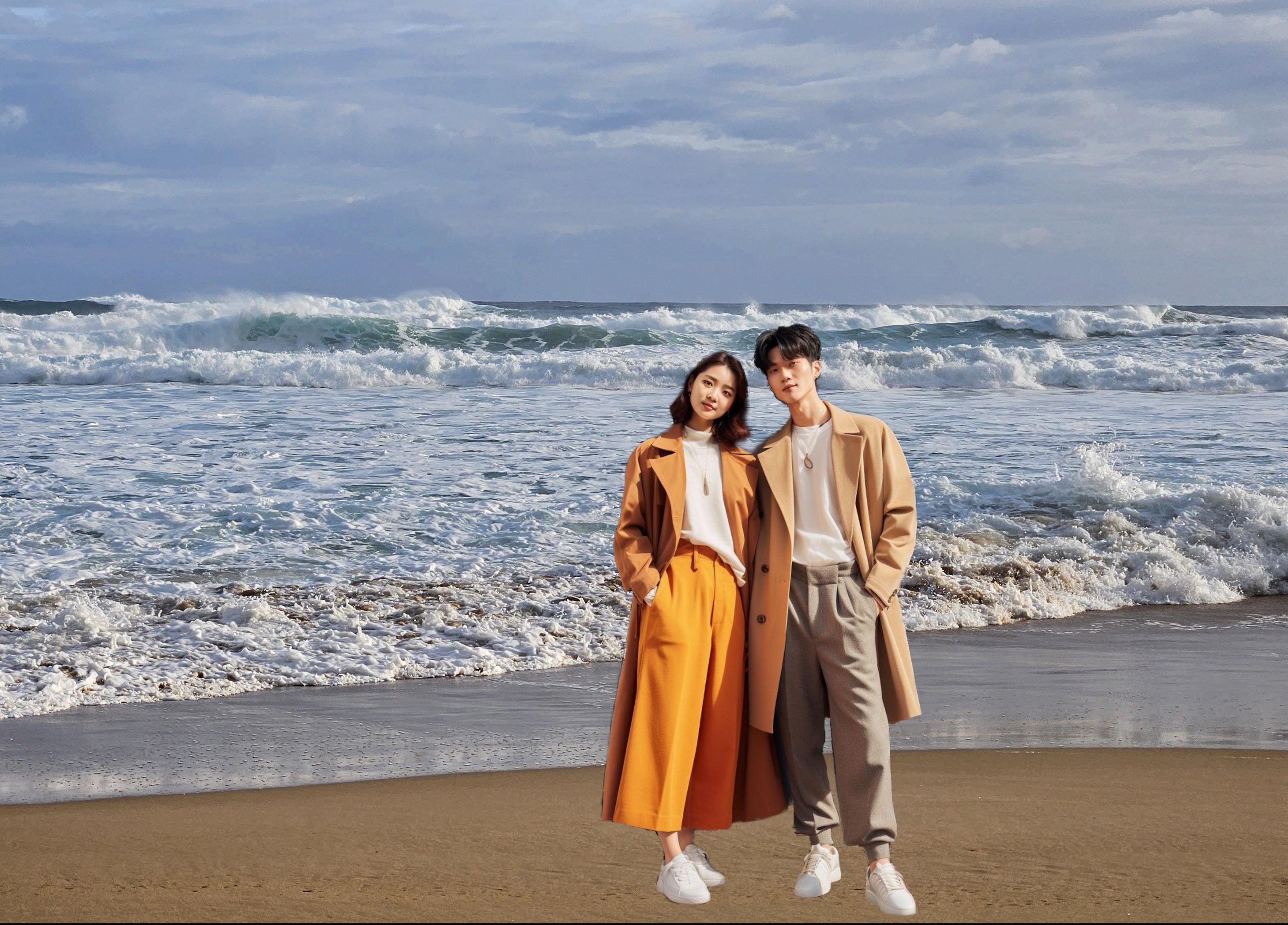}
        \hfill
        \includegraphics[height=2.7cm]{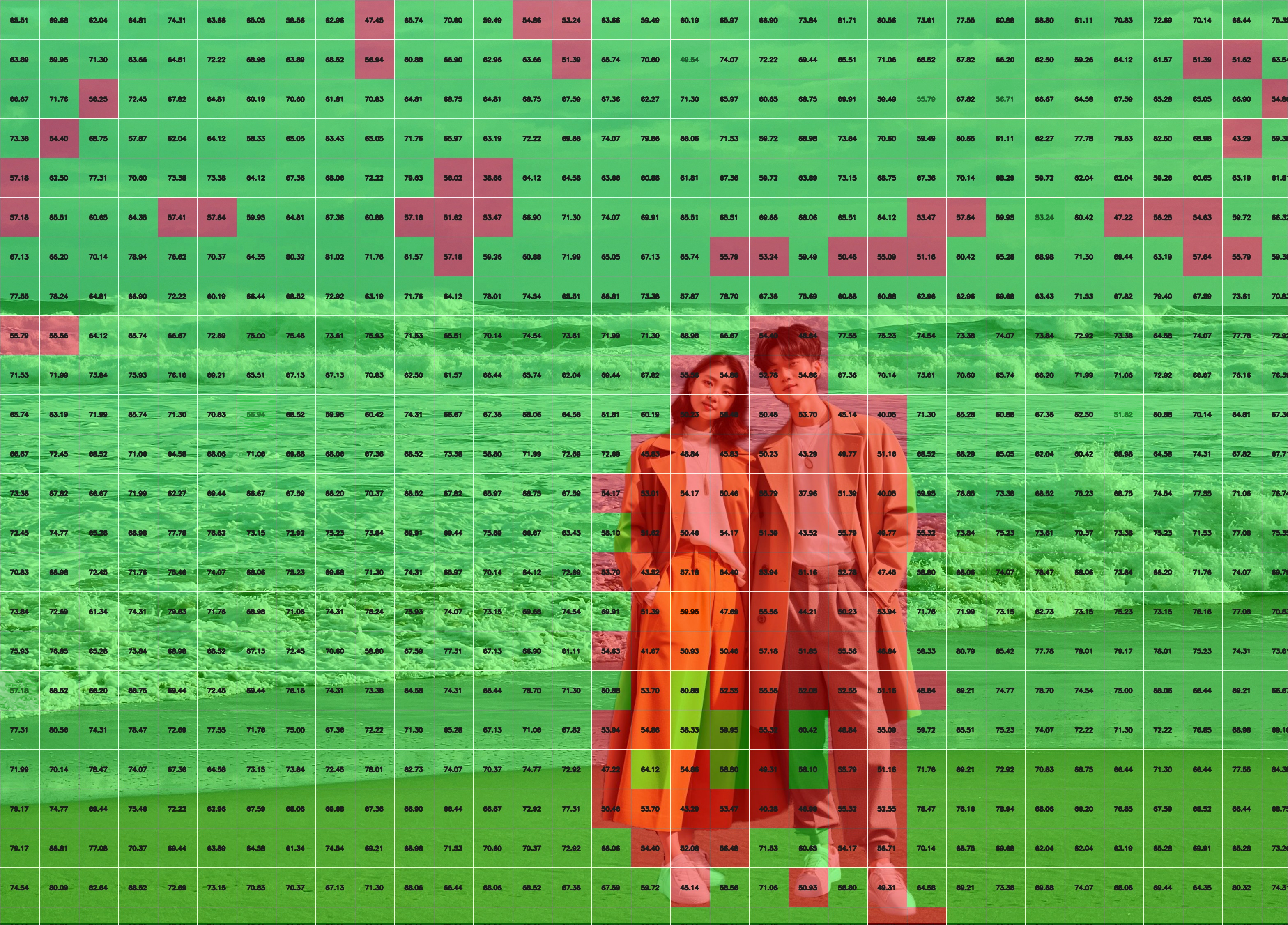}
        \hfill
        \null
        \caption{Watermark detection process on an image that has been manipulated post-watermarking.} \label{people} %
    \end{subfigure}
    \caption{When editing new content into a watermarked image, the manipulation becomes immediately detectable. While some red squares (false negatives) are visible in the watermarked image (\cref{watermarked_beach}), they are evenly scattered throughout the image rather than forming dense clusters. In \cref{people} we inserted two people into the watermarked image from \cref{watermarked_beach}. The inserted content, which does not contain the watermark, appears as a large red cluster in the detection map, clearly highlighting the area of image modification.
}  \label{fig:inserted_couple}
\end{figure}

\paragraph{Example: Inserted Non-watermarked Content.}
Conversely, \cref{fig:inserted_couple} demonstrates a scenario where non-watermarked content has been incorporated into a watermarked image. The detection map shows a prominent red cluster (corresponding to the inserted couple) within a predominantly green area. This demonstrates the utility of our localised approach for digital forensics and helps highlight the impact of manipulations on the distributed watermark.

\section{Experiments and Analysis}
\label{sec:eval}

In this section, we evaluate our watermarking scheme's performance, focusing on its detectability, interpretability for image forensics, and robustness against common image manipulations. We begin by identifying a set of image manipulations that preserve visual quality, then assess our method across multiple datasets.  We then assess watermark detectability across multiple datasets, examine our scheme's interpretability for image forensics, and evaluate its robustness against common image corruptions. The parameter settings used for our method are detailed in App.~\ref{app:embedding_settings}.

\subsection{Experimental Setup}

\subsubsection{Datasets}

We employ two real-world datasets: MS-COCO~\cite{coco}, where we select the first 5,000 images from the test set and WikiArt~\cite{wikiart} with the first 1,000 images. Additionally, we utilise two AI-generated datasets: 3,310 images generated by DALL-E~3~\cite{dalle_paper}, accessed through Hugging Face Datasets~\cite{dalle3_webpage} and 5,000 images from the DiffusionDB dataset~\cite{diffdb}. This selection ensures that experiments cover a broad range of image resolutions. A distribution of image sizes for each dataset is provided in \cref{fig:pixel_boxplot} in App.~\ref{app:image_sizes}.

\subsubsection{Evaluation Metrics}

For image quality assessment, we utilise three common metrics to compare watermarked images with their original counterparts: \textbf{PSNR}, where higher values indicate better image quality; \textbf{SSIM}, for which higher scores reflect greater structural similarity to the original image and \textbf{LPIPS}, where lower scores indicate closer structural similarity.

To evaluate the robustness of watermarking schemes, we employ two key metrics. \textbf{Watermark detection rate} (WDR), equivalent to the true positive rate (TPR), measures the proportion of watermarked images correctly identified as watermarked. It ranges from 0 to 1, with higher values indicating better performance. 
\textbf{False positive rate} (FPR) assesses the proportion of non-watermarked images incorrectly identified as watermarked. It also ranges from 0 to 1, with lower values indicating better performance.

\subsubsection{Image Manipulations} \label{sec:attack_methods}

To assess the robustness of our method, we consider a range of image manipulations, building on the comprehensive evaluation provided by \cite{zodiac}. These include:
\begin{itemize}
    \item \textbf{Brightness} and \textbf{contrast} modifications with factor 0.5
    \item \textbf{JPEG compression}, using a quality setting of 50
    \item Adding \textbf{Gaussian noise} with a standard deviation of 0.05
    \item \textbf{Gaussian blur} with a kernel size of 5 and $\sigma=1$ 
    \item \textbf{Rotating} the image by 90 degrees
    \item \textbf{VAE-based image compression models}, Bmshj18 \cite{balle2018} and Cheng20 \cite{cheng2020}, with a quality level of 3.
\end{itemize}

Our primary focus is on transformations that preserve visual quality. Some of these modifications, particularly the VAE-based compression models (Cheng20 and Bmshj18) and extreme brightness adjustments, can significantly alter the appearance of the image as shown in \cref{fig:attacks_vis} of App.~\ref{app:attack-image-quality}. 
While we include these in our experiments for completeness and comparison with prior work, we consider them outside our primary scope of evaluation.

\begin{figure*}[h!]
    \centering
    \begin{minipage}{0.71\textwidth}
        \centering
        \begin{subfigure}{0.32\textwidth}
            \centering
            \includegraphics[width=\textwidth]{images/unsplash_DWT_cs_8_diag_2_dwt_lvl_3.jpg}
            \caption{Watermarked image \cite{unsplash_image}}
        \end{subfigure}
        \hfill
        \begin{subfigure}{0.32\textwidth}
            \centering
            \includegraphics[width=\textwidth]{images/unsplash_edited.jpg}
            \caption{Edited watermarked image}
        \end{subfigure}
        \hfill
        \begin{subfigure}{0.32\textwidth}
            \centering
            \includegraphics[width=\textwidth]{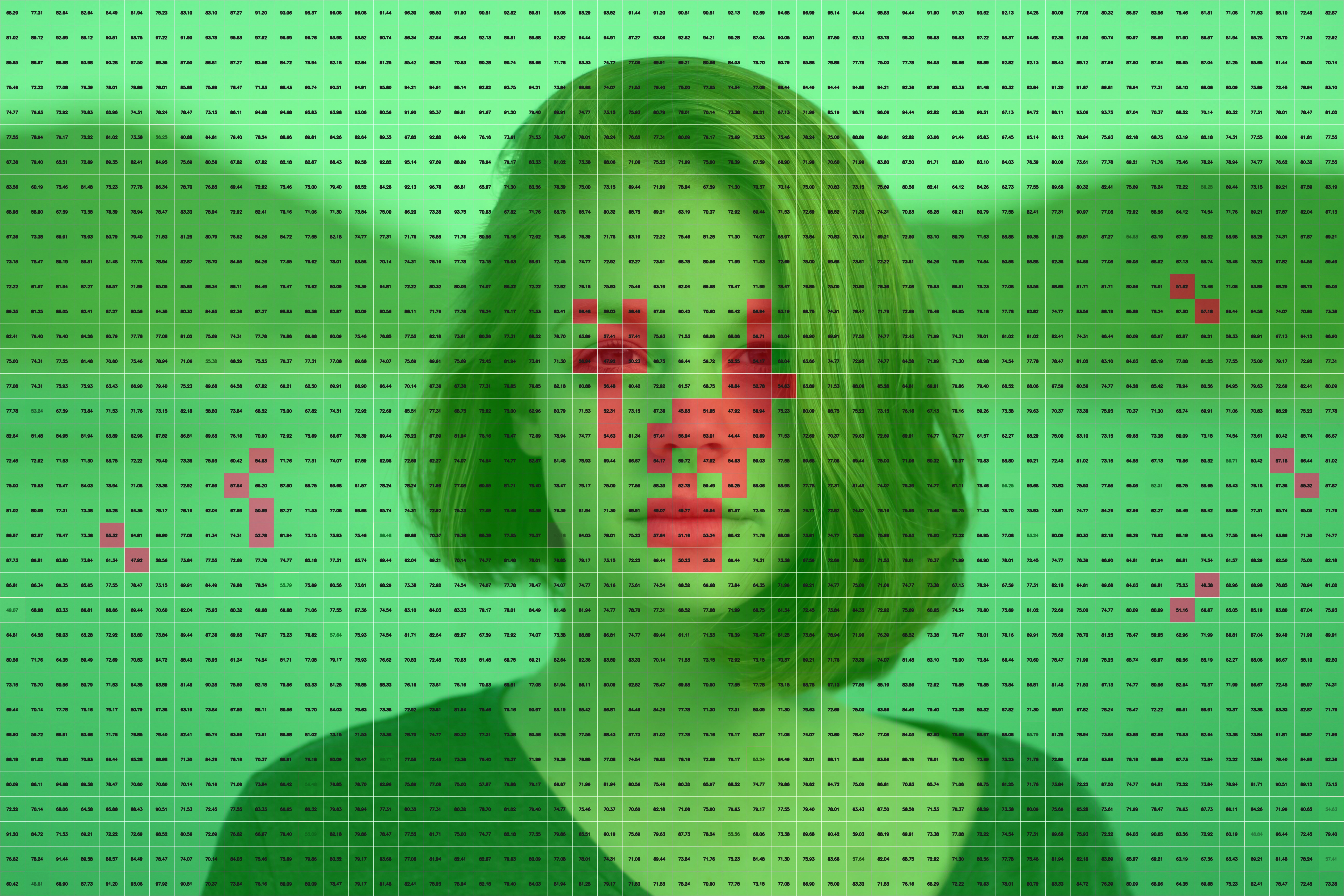}
            \caption{Detection map}
        \end{subfigure}
        \caption{Our watermark demonstrates high sensitivity to image warping operations, enabling the detection of subtle manipulations.}
        \label{fig:face_warp}
    \end{minipage}%
    \hfill
    \begin{minipage}{0.26\textwidth}
        \centering
        \renewcommand{\arraystretch}{1.2}
        \vspace{1.05em} %
        \resizebox{\linewidth}{!}{%
          \begin{tabular}{|l|c|c|}
  \hline
  \rowcolor{gray!30}
  \textbf{Dataset} & \textbf{Orig.} & \textbf{Waterm.} \\
  \hline
  COCO       & 18.62\% & \cellcolor{YellowGreen!40}92.51\% \\
  DALL-E     & 16.82\% & \cellcolor{YellowGreen!40}95.24\% \\
  DiffDB& 18.46\% & \cellcolor{YellowGreen!40}92.31\% \\
  WikiArt    & 19.49\% & \cellcolor{YellowGreen!40}93.33\% \\
  \hline
\end{tabular}

}
        \vspace{1.05em}
        \captionof{table}{Watermarked area detected in original vs.\ watermarked images.}
        \label{tab:detectability}
    \end{minipage}
\end{figure*}

\subsection{Detectability} \label{sec:exp_detect}

To demonstrate the effectiveness of our watermark detection across different image types, we analyse the scores generated for both original and watermarked images. 
In this experiment, we embed the watermark with an interval length of $l=8$ and conduct hypothesis tests using a 5\% significance level for each block. We then calculate the percentage of watermarked blocks per image. The results are presented in \cref{tab:detectability}.
Our findings show a clear distinction between watermarked and non-watermarked images. For watermarked images, the percentage of detected watermarked area exceeds 90\%, while for non-watermarked images, this percentage remains below 20\%. This provides strong evidence that our watermark is reliably detectable. 
We additionally compute FPR and ROC AUC, where we observe perfect detection scores of 0\% and 1 across all datasets.

\begin{table*}[h!]
\centering
\scriptsize
\resizebox{\textwidth}{!}{%
\begin{tabular}{lcccccccccccc}
\hline
Method      & PSNR $\uparrow$  & SSIM $\uparrow$    &  LPIPS $\downarrow$    & Pre-Attack & Rotation & JPEG & Blur & Contrast & Noise & Cheng20 & Brightness & Bmshj18 \\ \hline
\multicolumn{13}{c}{\textbf{MS-COCO}} \\ \hline

\textbf{MELB}       & 36.07 & \cellcolor{YellowGreen!40} 0.99 & \cellcolor{YellowGreen!40} 0.007 & \cellcolor{MidnightBlue!35}1.000 & \cellcolor{MidnightBlue!35}0.973 & \cellcolor{MidnightBlue!35}0.969 & \cellcolor{MidnightBlue!35}0.991 & 0.160 & \cellcolor{MidnightBlue!20}0.877 & \textcolor{gray}{0.020} & \textcolor{gray}{0.606} & \textcolor{gray}{0.041} \\
DwtDct \cite{cox}      & 37.88 & \cellcolor{YellowGreen!20}0.97 & 0.02 & \cellcolor{MidnightBlue!10}0.790 & 0.000 & 0.000 & 0.156 & 0.000 & 0.687 & \textcolor{gray}{0.000} & \textcolor{gray}{0.000} & \textcolor{gray}{0.000} \\
DwtDctSvd \cite{cox}  & 38.06 & \cellcolor{YellowGreen!20}0.98 & 0.02 & \cellcolor{MidnightBlue!35}1.000 & 0.000 & \cellcolor{MidnightBlue!10}0.746 & \cellcolor{MidnightBlue!35}1.000 & 0.100 & \cellcolor{MidnightBlue!35}0.998 & \textcolor{gray}{0.032} & \textcolor{gray}{0.098} & \textcolor{gray}{0.016} \\
RivaGAN \cite{rivagan}    & 40.57 & \cellcolor{YellowGreen!20}0.98 & 0.04 & \cellcolor{MidnightBlue!35}1.000 & 0.000 & \cellcolor{MidnightBlue!35}0.984 & \cellcolor{MidnightBlue!35}1.000 & \cellcolor{MidnightBlue!35}0.998 & \cellcolor{MidnightBlue!35}1.000 & \textcolor{gray}{0.010} & \cellcolor{gray!30}\textcolor{gray}{0.996} & \textcolor{gray}{0.010} \\
SSL  \cite{SSL}       & \cellcolor{YellowGreen!40}41.81 & \cellcolor{YellowGreen!20}0.98 & 0.06 & \cellcolor{MidnightBlue!35}1.000 & \cellcolor{MidnightBlue!35}0.952 & 0.046 & \cellcolor{MidnightBlue!35}1.000 & \cellcolor{MidnightBlue!35}0.996 & 0.038 & \textcolor{gray}{0.000} & \cellcolor{gray!30}\textcolor{gray}{0.992} & \textcolor{gray}{0.000} \\
CIN  \cite{CIN}       & \cellcolor{YellowGreen!20}41.77 & \cellcolor{YellowGreen!20}0.98 & 0.02 & \cellcolor{MidnightBlue!35}1.000 & 0.216 & \cellcolor{MidnightBlue!20}0.944 & \cellcolor{MidnightBlue!35}1.000 & \cellcolor{MidnightBlue!35}1.000 & \cellcolor{MidnightBlue!35}1.000 & \textcolor{gray}{0.666} & \cellcolor{gray!30}\textcolor{gray}{1.000} & \textcolor{gray}{0.662} \\
StegaStamp \cite{stegastamp} & 28.64 & 0.91 & 0.13 & \cellcolor{MidnightBlue!35}1.000 & 0.000 & \cellcolor{MidnightBlue!35}1.000 & \cellcolor{MidnightBlue!35}1.000 & \cellcolor{MidnightBlue!35}0.998 & \cellcolor{MidnightBlue!35}0.998 & \cellcolor{gray!30}\textcolor{gray}{1.000} & \cellcolor{gray!30}\textcolor{gray}{0.998} & \cellcolor{gray!30}\textcolor{gray}{0.998} \\
ZoDiac \cite{zodiac}     & 29.41 & 0.92 & 0.09 & \cellcolor{MidnightBlue!35}0.998 & 0.538 & \cellcolor{MidnightBlue!35}0.992 & \cellcolor{MidnightBlue!35}0.996 & \cellcolor{MidnightBlue!35}0.998 & \cellcolor{MidnightBlue!35}0.996 & \cellcolor{gray!30}\textcolor{gray}{0.986} & \cellcolor{gray!30}\textcolor{gray}{0.998} & \cellcolor{gray!30}\textcolor{gray}{0.996} \\ \hline

\multicolumn{13}{c}{\textbf{DiffDB}} \\ \hline

\textbf{MELB}       & \cellcolor{YellowGreen!40}43.45 & \cellcolor{YellowGreen!20}0.97 & 0.03 & \cellcolor{MidnightBlue!35}1.000 & \cellcolor{MidnightBlue!35}0.973 & \cellcolor{MidnightBlue!35}0.961 & \cellcolor{MidnightBlue!35}0.979 & 0.091 & \cellcolor{MidnightBlue!20}0.882 & \textcolor{gray}{0.020} & \textcolor{gray}{0.570} & \textcolor{gray}{0.032} \\
DwtDct      & 37.77 & \cellcolor{YellowGreen!20}0.96 & \cellcolor{YellowGreen!40}0.02 & 0.690 & 0.000 & 0.000 & 0.224 & 0.000 & 0.574 & \textcolor{gray}{0.000} & \textcolor{gray}{0.000} & \textcolor{gray}{0.000} \\
DwtDctSvd   & 37.84 & \cellcolor{YellowGreen!20}0.97 & \cellcolor{YellowGreen!40}0.02 & \cellcolor{MidnightBlue!35}0.998 & 0.000 & \cellcolor{MidnightBlue!10}0.812 & \cellcolor{MidnightBlue!35}0.996 & 0.088 & \cellcolor{MidnightBlue!35}0.982 & \textcolor{gray}{0.030} & \textcolor{gray}{0.088} & \textcolor{gray}{0.014} \\
RivaGAN     & 40.60 & \cellcolor{YellowGreen!40}0.98 & 0.04 & \cellcolor{MidnightBlue!35}0.974 & 0.000 & \cellcolor{MidnightBlue!20}0.898 & \cellcolor{MidnightBlue!35}0.966 & \cellcolor{MidnightBlue!20}0.932 & \cellcolor{MidnightBlue!35}0.958 & \textcolor{gray}{0.004} & \cellcolor{gray!20}\textcolor{gray}{0.932} & \textcolor{gray}{0.008} \\
SSL         & 41.84 & \cellcolor{YellowGreen!40}0.98 & 0.06 & \cellcolor{MidnightBlue!35}0.998 & \cellcolor{MidnightBlue!20}0.898 & 0.040 & \cellcolor{MidnightBlue!35}1.000 & \cellcolor{MidnightBlue!35}0.996 & 0.030 & \textcolor{gray}{0.000} & \cellcolor{gray!30}\textcolor{gray}{0.990} & \textcolor{gray}{0.000} \\
CIN         & 39.99 & \cellcolor{YellowGreen!40}0.98 & \cellcolor{YellowGreen!40}0.02  & \cellcolor{MidnightBlue!35}1.000 & 0.212 & \cellcolor{MidnightBlue!20}0.942 & \cellcolor{MidnightBlue!35}0.998 & \cellcolor{MidnightBlue!35}1.000 & \cellcolor{MidnightBlue!35}0.998 & \textcolor{gray}{0.478} & \cellcolor{gray!30}\textcolor{gray}{1.000} & \textcolor{gray}{0.662} \\
StegaStamp  & 28.51 & 0.90 & 0.13 & \cellcolor{MidnightBlue!35}1.000 & 0.000 & \cellcolor{MidnightBlue!35}1.000 & \cellcolor{MidnightBlue!35}0.998 & \cellcolor{MidnightBlue!35}0.998 & \cellcolor{MidnightBlue!35}0.998 & \textcolor{gray}{0.286} & \cellcolor{gray!30}\textcolor{gray}{0.998} & \cellcolor{gray!30}\textcolor{gray}{0.996} \\
ZoDiac      & 29.18 & 0.92 & 0.07 & \cellcolor{MidnightBlue!35}1.000 & 0.558 & \cellcolor{MidnightBlue!35}0.994 & \cellcolor{MidnightBlue!35}1.000 & \cellcolor{MidnightBlue!35}0.998 & \cellcolor{MidnightBlue!35}0.998 & \cellcolor{gray!30}\textcolor{gray}{0.988} & \cellcolor{gray!30}\textcolor{gray}{0.998} & \cellcolor{gray!30}\textcolor{gray}{0.994} \\ \hline

\multicolumn{13}{c}{\textbf{WikiArt}} \\ \hline

\textbf{MELB}       & 30.44 & \cellcolor{YellowGreen!40}0.99 & \cellcolor{YellowGreen!40}0.01 & \cellcolor{MidnightBlue!35}1.000 & \cellcolor{MidnightBlue!35}0.980 & \cellcolor{MidnightBlue!35}0.984 & \cellcolor{MidnightBlue!35}0.970 & 0.241 & \cellcolor{MidnightBlue!20}0.859 & \textcolor{gray}{0.020} & \textcolor{gray}{0.470} & \textcolor{gray}{0.020} \\
DwtDct      & 38.84 & \cellcolor{YellowGreen!20}0.97 & 0.02 & \cellcolor{MidnightBlue!10}0.754 & 0.000 & 0.000 & 0.280 & 0.000 & 0.594 & \textcolor{gray}{0.000} & \textcolor{gray}{0.000} & \textcolor{gray}{0.000} \\
DwtDctSvd   & 39.14 & \cellcolor{YellowGreen!20}0.98 & 0.02 & \cellcolor{MidnightBlue!35}1.000 & 0.000 & 0.698 & \cellcolor{MidnightBlue!35}1.000 & 0.094 & \cellcolor{MidnightBlue!35}1.000 & \textcolor{gray}{0.000} & \textcolor{gray}{0.096} & \textcolor{gray}{0.034} \\
RivaGAN     & 40.44 & \cellcolor{YellowGreen!20}0.98 & 0.05 & \cellcolor{MidnightBlue!35}1.000 & 0.000 & \cellcolor{MidnightBlue!20}0.990 & \cellcolor{MidnightBlue!35}1.000 & \cellcolor{MidnightBlue!20}1.000 & \cellcolor{MidnightBlue!35}1.000 & \textcolor{gray}{0.072} & \cellcolor{gray!30}\textcolor{gray}{0.998} & \textcolor{gray}{0.010} \\
SSL         & \cellcolor{YellowGreen!20}41.81 & \cellcolor{YellowGreen!20}0.99 & 0.06 & \cellcolor{MidnightBlue!35}1.000 & \cellcolor{MidnightBlue!20}0.932 & 0.082 & \cellcolor{MidnightBlue!35}1.000 & \cellcolor{MidnightBlue!35}0.998 & 0.108 & \textcolor{gray}{0.000} & \cellcolor{gray!30}\textcolor{gray}{0.988} & \textcolor{gray}{0.000} \\
CIN         & \cellcolor{YellowGreen!40}41.92 & \cellcolor{YellowGreen!20}0.98 & 0.02 & \cellcolor{MidnightBlue!35}1.000 & 0.228 & \cellcolor{MidnightBlue!20}0.936 & \cellcolor{MidnightBlue!35}1.000 & \cellcolor{MidnightBlue!35}1.000 & \cellcolor{MidnightBlue!35}0.998 & \textcolor{gray}{0.664} & \cellcolor{gray!30}\textcolor{gray}{1.000} & \textcolor{gray}{0.664} \\
StegaStamp  & 28.45 & 0.91 & 0.14 & \cellcolor{MidnightBlue!35}1.000 & 0.000 & \cellcolor{MidnightBlue!35}1.000 & \cellcolor{MidnightBlue!35}1.000 & \cellcolor{MidnightBlue!35}0.998 & \cellcolor{MidnightBlue!35}1.000 & \cellcolor{gray!30}\textcolor{gray}{1.000} & \cellcolor{gray!30}\textcolor{gray}{1.000} & \cellcolor{gray!30}\textcolor{gray}{1.000} \\
ZoDiac      & 30.04 & 0.92 & 0.10 & \cellcolor{MidnightBlue!35}1.000 & 0.478 & \cellcolor{MidnightBlue!35}1.000 & \cellcolor{MidnightBlue!35}1.000 & \cellcolor{MidnightBlue!35}1.000 & \cellcolor{MidnightBlue!35}0.998 & \cellcolor{gray!30}\textcolor{gray}{1.000} & \cellcolor{gray!30}\textcolor{gray}{1.000} & \cellcolor{gray!30}\textcolor{gray}{0.994} \\ \hline

\cellcolor{gray!30}\textbf{Colour keys:} & Norms: & \cellcolor{YellowGreen!40}Best Value & \cellcolor{YellowGreen!20}within 2\% & & WDR: & $<$ 0.70 & \cellcolor{MidnightBlue!10} 0.70 - 0.85 & \cellcolor{MidnightBlue!20} 0.85 - 0.95 & \cellcolor{MidnightBlue!35} $>$0.95 & & & \\ \hline
\end{tabular}
}
\caption{Comparison of watermarking schemes according to perceptibility (PSNR, SSIM, LPIPS) and robustness to image manipulations (WDR post-manipulation). A higher WDR post-manipulation means the watermarking scheme is more robust. Grey-shaded manipulations are considered out-of-scope due to their impact on image quality.}
\label{tab:comparison}
\end{table*}

\subsection{Image Forensics} \label{sec:im_forensics}

Our method has potential applications in image forensics owing to its ability to produce a localised detection map. As shown in \cref{sec:Map}, this map can highlight areas that may have been modified or inserted post-watermarking by analysing the distribution and clustering of red and green blocks. The same principle applies to subtle image manipulations, such as shape changes using common warp tools.

In \cref{fig:face_warp}, we present an example image of a woman with applied face modifications such as making the eyes and lips bigger, and slightly changing the shape of the nose. Our localised detection map clearly highlights the correct areas of modification.
While our watermark is robust to noise or compression, which apply uniformly across the image and primarily affect higher frequency components, even small warping transformations can break it. This sensitivity occurs because we embed our watermark in the LL band of the DWT transform. The DWT decomposes the image into different frequency subbands, with the LL band representing the large-scale approximation of the image. This band contains the most significant information about the edges and shapes, and is sensitive to geometric transformations.

\subsection{Robustness} \label{sec:robustness}

\subsubsection{Cropping} \label{sec:exp_cropping}

We evaluate our method for different cropping strengths and present the results in \cref{fig:cropping_results}. Since we are using the brute-force detection algorithm presented in \cref{sec:cropping}, we work with a subsample of 1000 images per dataset. We set the stopping criterion to $p=0.8$, which yields a FPR of $\approx 0.01$ for each dataset. A cropping strength of \eg, 70\% means that we randomly remove 30\% of the image area. Our results indicate that our method is highly robust against cropping, maintaining a TPR of $>90\%$ for all datasets when up to 30\% of the image area is cropped away. Even under radical cropping conditions where half of the image is removed, our method still achieves high TPR values ranging between 76\% and 91\%. While the performance on all datasets is strong, the results for WikiArt and DALL-E stand out as particularly robust. This can be attributed to their larger image sizes, providing more blocks for evaluation. A higher number of blocks leads to more stable detection statistics and reduces the influence of random variation.

\subsubsection{Other Image Manipulations} \label{sec:geom_attacks}

In \cref{tab:comparison}, we compare the performance of our method with other post-hoc watermarking approaches, based on results reported in \cite{zodiac}. Across all similarity metrics, our method achieves high imperceptibility---producing watermarked images that closely resemble the originals.

In terms of robustness, our method performs comparably to existing techniques when evaluated against common image transformations such as JPEG compression, Gaussian noise, and blur. While certain watermarks outperform ours in overall robustness, particularly under extreme perturbations, our method remains sufficiently resilient to modifications that do not substantially degrade image quality.  Our approach is less robust to contrast and brightness adjustments. However, this aligns with our threat model, which assumes a malicious actor is unlikely to apply transformations that visibly distort the image. \Eg, a brightness factor of 0.5 darkens the image by 50\%, resulting in a perceptually significant alteration. Since our method is designed to detect tampering that preserves visual plausibility, we consider this trade-off acceptable within our intended use case.

\section{Conclusion}
\label{sec:conclusion}

We propose a simple yet effective post-hoc image watermarking method, \melb, that combines localised embedding with region-level interpretability. When embedding, our method operates in the discrete wavelet transform domain, using a block-wise approach that randomly divides the transform coefficients into allowed and disallowed regions, then projects the image blocks onto the nearest allowed regions. For detection, we adopt a one-sided hypothesis test to determine whether an image is watermarked.
Our method outputs not only an overall detection score for the whole image but also a detection map, highlighting the areas where the watermark is found, providing users more transparency. %
Compared to prior post-hoc methods, \melb\ offers more interpretable detection while retaining competitive robustness against rotation, JPEG compression, blur, and noise, all while introducing minimal visible artefacts. Notably, our method shows strong robustness to cropping, maintaining detectability even when up to half of the image is cropped away.
Future work could explore alternative approaches to seed computation beyond our current method of using the rounded mean colour of a block, \eg, investigating various hashing functions.

\section*{Acknowledgements}

This research was supported by The University of Melbourne’s Research Computing Services and the Petascale Campus Initiative. We also acknowledge the assistance of ChatGPT and GitHub Copilot for scaffolding code in the released supplementary material.

{
    \small
    \bibliographystyle{ieeenat_fullname}
    \bibliography{main}
}

\clearpage
\appendix
\section{Further Background}

\subsection{Discrete Wavelet Transform} \label{app:dwt}

The Discrete Wavelet Transform (DWT) is a method that converts an image from its spatial representation to a frequency-based representation. When applied to an image, the DWT creates four frequency subbands: low-low (LL), low-high (LH), high-low (HL), and high-high (HH). To achieve additional levels of DWT, each subband can be further decomposed into four new subbands.
The LL subband, containing the lowest frequency components, holds most of the image's weight and important structural information. While it is resilient to subtle, image-wide changes such as noise and compression, it demands precise editing to preserve image quality, as alterations in this band are more likely to create visible artefacts \cite{buch}.
Higher frequency subbands (LH, HL, and HH) capture fine details and edges of the image. These components are more easily affected by typical image modifications, such as JPEG compression.
To embed a watermark, we carefully alter the coefficients in the LL subband. After embedding, we can reconstruct the original image using the inverse DWT.

\subsection{Entropy of Visual Data} \label{app:entropy}

Image entropy, specifically Shannon entropy, quantifies the randomness or information content of an image~\cite{shannon_entropy}. It is calculated from the empirical probability distribution of pixel intensities. For a grayscale image, entropy $H$ is calculated as 
\begin{align}
    H = -\sum p(i)\log(p(i)), \label{entropy_eq}
\end{align} 
where $p(i)$ represents the frequency of intensity level $i$ in the image. Intuitively, regions with high entropy signify areas in the image where there is a lot of variation in pixel intensities, such as edges, textures, and details. These complex regions are often more suitable for embedding watermarks because the changes introduced by the watermark are less likely to be noticeable to the human eye \cite{entropy_based_wm}.

\section{Advanced Considerations and Variants of \melb} \label{app:melb}

This section provides the core algorithmic components behind our watermarking method. We provide pseudocode and formal descriptions of the embedding and detection procedures and formulate the statistical hypothesis test used for detection. We also present an optional extension of our method that strengthens the system against brute-force attacks and elaborate on the details of two further refinements that were used in our experiments in \cref{sec:robustness}.

\subsection{Algorithms} \label{app:algorithms}

In this section, we present additional algorithms for the detection process, as described in \cref{sec:embedding} and \cref{sec:Detection}. \cref{alg:MELB} and \cref{alg:MELB_detection} represent the main detection steps, while \cref{alg:eval} summarises the evaluation procedure. The statistical hypothesis test refers to the procedure described in \cref{sec:Detection}, where we determine whether a block is watermarked by assessing the number of its DWT coefficients that fall within the allowed domain. The optional $\mathsf{post\_process}$ step, introduced in \cref{sec:Map}, applies a low-pass filter to the detection map to enhance its utility as a visual aid. This post-processing step does not influence the final image-level score.

\begin{algorithm}
\caption{Embedding Algorithm for \melb}\label{alg:MELB}
\small
\begin{algorithmic}[1]
    \Require Image $I$, interval length $l$, block size $k$, sub-block size $m$, number of DWT-levels $d$, DWT coefficient limit $r$, key $\sk$
    \algrule
    \Procedure{$\embed$}{$I$, $l$, $k$, $m$, $d$, $r$}
        \State Decompose image into channels $R, G, B \gets I_{:, :, 0}, I_{:, :, 1}, I_{:, :, 2}$
        \For{each channel $C \in \{R, G, B\}$}
            \State Partition $C$ into non-overlapping $m \times m$ blocks 
            \For{each block $b$}
                \State $\seed \gets \getSeed(b)$ 
                \State $b^\star \gets \embedBlock(b, \seed, k, d, \sk)$
                \State Replace block $b$ in $C$ with watermarked block $b^\star$
            \EndFor
        \EndFor
        \State \textbf{return} Image made of watermarked $R, G, B$ channels 
    \EndProcedure
\\
    \Procedure{$\embedBlock$}{$b$, $seed$, $k$, $d$, $l$, $r$, $\sk$}
        \State Use $\seed$ to generate a binary sequence $S \gets F_{\sk}(\seed)$ \label{alg:MELB:binary-seq}
        \State Split the DWT domain interval $[-r, r]$ into sub-intervals 
        \State of length $l$
        \State Generate red list $R$ and green list $G$ by colouring the 
        \State intervals based on the sequence $S$
        \State Partition $b$ into $k \times k$ sub-blocks
        \For{each sub-block $s \in b$}
            \State Apply $d$ levels of DWT to $s$ and extract LL coefficients
            \For{each coefficient $c$ in $s$}
                \If{$c \in R$}
                    \State $c^\star \gets \perturb(c, G)$
                \EndIf
            \EndFor
        \EndFor
        \State \textbf{return} Watermarked block $b^\star$
    \EndProcedure
\\
    \Procedure{$\perturb$}{$c$, $G$}
        \State Find green interval $[g_{\text{min}}, g_{\text{max}}] \in G$ that is closest to $c$
    \State $\tilde c \gets (g_{\text{min}} + g_{\text{max}})/2$ 
    \State \textbf{return} $\tilde c$
    \EndProcedure
\end{algorithmic}
\end{algorithm}

\begin{algorithm}
\caption{Evaluation Procedure}\label{alg:eval}
\small
\begin{algorithmic}[1]
\Procedure{$\eval$}{$\mathcal{S}$, $\alpha$}
    \State Initialise detection map $\mathcal{M}$ with same shape as $\mathcal{S}$
    \State Initialise green count $n \gets 0$
    \For{each position $(i, j)$ in $\mathcal{S}$}
    \State \textcolor{gray}{\Comment{$\mathcal{S}[i,j]$ stores the number of watermarked coefficients}}
        \State Perform one-sided hypothesis test on $\mathcal{S}[i,j]$ at 
        \State significance level $\alpha$
        \If{$H_0$ is rejected}
            \State $\mathcal{M}[i,j] \gets$ green
            \State $n \gets n + 1$
        \Else
            \State $\mathcal{M}[i,j] \gets$ red
        \EndIf
    \EndFor
    \State $\mathsf{post\_process}(\mathcal{M})$ \textcolor{gray}{\Comment{optional low-pass filter}}
    \State Compute $\score \gets \frac{n}{|\mathcal{S}|}$
    \State \textbf{return} $\score$, $\mathcal{M}$
\EndProcedure
\end{algorithmic}
\end{algorithm}

\begin{algorithm}
\caption{Calculating the Seed}\label{alg:get_seed}
\small
\begin{algorithmic}[1]
\Procedure{$\getSeed$}{$s$, $\roundVal \gets 30$}
    \State Compute the mean pixel intensity $\mu$ of all pixels in $s$ 
    \State Round $\mu$ to the nearest multiple of $\roundVal$
    \State \textbf{return} rounded value 
\EndProcedure
\end{algorithmic}
\end{algorithm}

\begin{algorithm}
\caption{Detection Algorithm for \melb}\label{alg:MELB_detection}
\small
\begin{algorithmic}[1]
    \Require Image $I$, interval length $l$, block size $k$, sub-block size $m$, number of DWT-levels $d$, significance level $\alpha$, key $\sk$
    \algrule
    \Procedure{$\detect$}{$I$, $l$, $k$, $m$, $d$, $\alpha$}
        \State Decompose image into channels $R, G, B \gets I_{:, :, 0}, I_{:, :, 1}, I_{:, :, 2}$
        \State Obtain score maps $\mathcal{S}_R, \mathcal{S}_G, \mathcal{S}_B$ by applying $\detectChannel$ 
        \State Compute mean score map $\mathcal{S}_{\text{mean}}$ over all channels
        \State $\score, \mathsf{map} \gets \eval(\mathcal{S}_{\text{mean}}, \alpha)$
        \State \textbf{return} $\score, \mathsf{map}$
    \EndProcedure
\\
\Procedure{$\detectChannel$}{$C$, $l$, $k$, $m$, $d$, $\alpha$}
    \State Partition $C$ into non-overlapping $m \times m$ blocks
    \State Initialise score matrix $\mathcal{S}$ with the same shape as block grid
    \For{each block $b$ at position $(i, j)$}
        \State $\seed \gets \getSeed(b)$
        \State $\mathcal{S}[i,j] \gets \detectBlock(b, \seed, k, d, \sk)$
    \EndFor
    \State \textbf{return} $\mathcal{S}$
\EndProcedure
\\
    \Procedure{$\detectBlock$}{$b$, $\seed$, $k$, $d$, $\sk$}
        \State Initialise $\countG \gets 0$ 
        \State Use $\seed$ to generate a binary sequence $S$ \label{alg:MELB_detection:binary-seq}
        \State Generate red list $R$ and green list $G$ according to $S$
        \State Partition $b$ into $k \times k$ sub-blocks
        \For{each sub-block $s \in b$}
            \State Apply $d$ levels of DWT to $s$ and extract LL coefficients~$\mathcal{L}$
            \State Increment $\countG$ by $|\mathcal{L} \cap G|$
        \EndFor
        \State \textbf{return} $\countG$ 
    \EndProcedure
    \end{algorithmic}
\end{algorithm}

\subsection{Parameter Configuration and Trade-offs} \label{app:parameters}
\begin{figure*}
    \centering
    \includegraphics[width=0.9\textwidth]{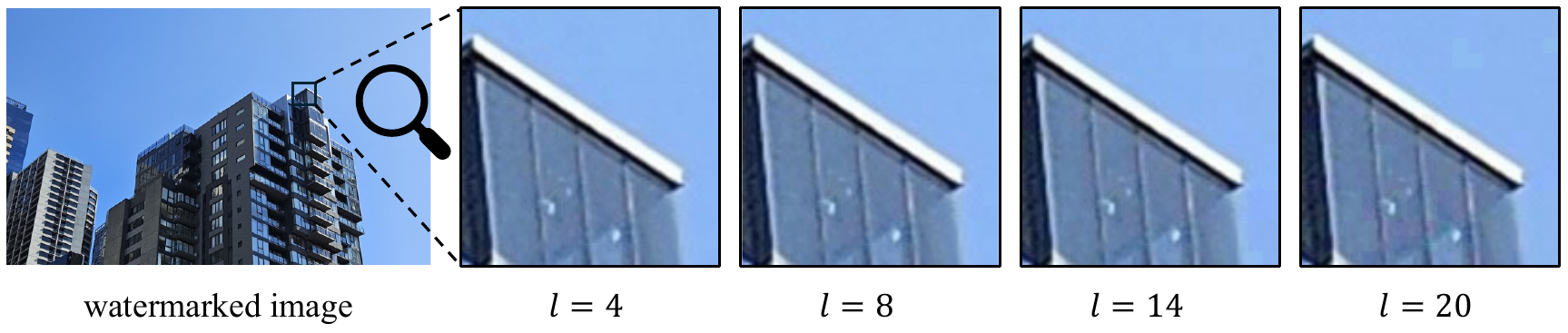} %
    \caption{Four watermarks using different interval lengths $l$. We find that using $l=8$ provides good image quality in all areas of the image. For larger interval lengths, we find that $l=14$ is the largest length not introducing visual artefacts in non-monochromatic regions. Excessively high values of $l$ lead to visible perturbations throughout the whole image. }
    \label{fig:diff_interval_lengths} %
\end{figure*}

The effectiveness of our watermarking method depends on multiple parameters. The most crucial parameter in our method is the \textbf{interval length} $l$, which controls the trade-off between watermark robustness and visual imperceptibility: shorter interval lengths result in less visible watermarks but decreased robustness, while longer intervals enhance robustness at the potential cost of visual perceptibility (see \cref{fig:diff_interval_lengths}). This is  explained by the fact that larger interval lengths provide greater tolerance for variations in the DWT coefficients, allowing them to undergo minor modifications while staying in the green domain. When choosing excessively large interval lengths, we suggest embedding the watermark only in high entropy areas to reduce the visual impact. %
It is practical to define a consistent \textbf{range} $[-r, r]$ that will be considered for the DWT coefficients that applies to all watermarked images. This ensures that the intervals are uniformly divided and always start from the same reference point, allowing the reliable reconstruction of the interval colouring during detection. While we can compute the theoretical bounds for DWT coefficients and work with the whole range, we note that most coefficients in natural images fall significantly below this bound. To reduce computational complexity and storage, we define the coefficient range as a parameter, chosen to cover the region where most DWT coefficients are expected to lie, and consider each value outside this range as allowed. 

We propose to define the $\getSeed$ function as the rounded mean colour of each image block. Generating distinct seeds per block is crucial, as it prevents uniform allowed value ranges across blocks, thereby strengthening security by avoiding repetitive patterns exploitable by attackers. Additionally, deriving each seed from the block's content enhances robustness to cropping, since the seed can be independently recovered even if parts of the image are removed. Using randomly generated seeds for each block would necessitate storing all seeds for detection, which is impractical at scale due to high storage overhead. Therefore, using the mean channel average strikes a practical balance between security, efficiency, and storage complexity. Exploring more sophisticated hash functions for seed generation is left as future work.

When the seed is derived from the rounded mean colour of the block, the \textbf{rounding value} presents a trade-off between the robustness against image manipulations and forgability. In general, a smaller rounding value would create more distinct seed values across the image, making it harder for malicious actors to gather sufficient samples of allowed and disallowed modifications for any particular seed value. This is critical because if too many blocks share the same seed value, an attacker could analyse the pattern of allowed and disallowed modifications across these blocks to approximate the watermarking rules. For instance, if multiple blocks with the same mean colour exhibit consistent patterns of allowed colour modifications, an attacker could learn and exploit these patterns to forge or remove the watermark. However, the rounding value cannot be too small, as it must also provide sufficient robustness against simple mean-shifting attacks. In \cref{sec:attack_seed} we demonstrate through experimental validation, that a rounding value of 30 achieves this balance effectively. Any attempt to shift block mean values enough to change their rounding behaviour results in severe degradation of visual quality.

\begin{figure*}
    \centering
    \includegraphics[width=0.9\textwidth]{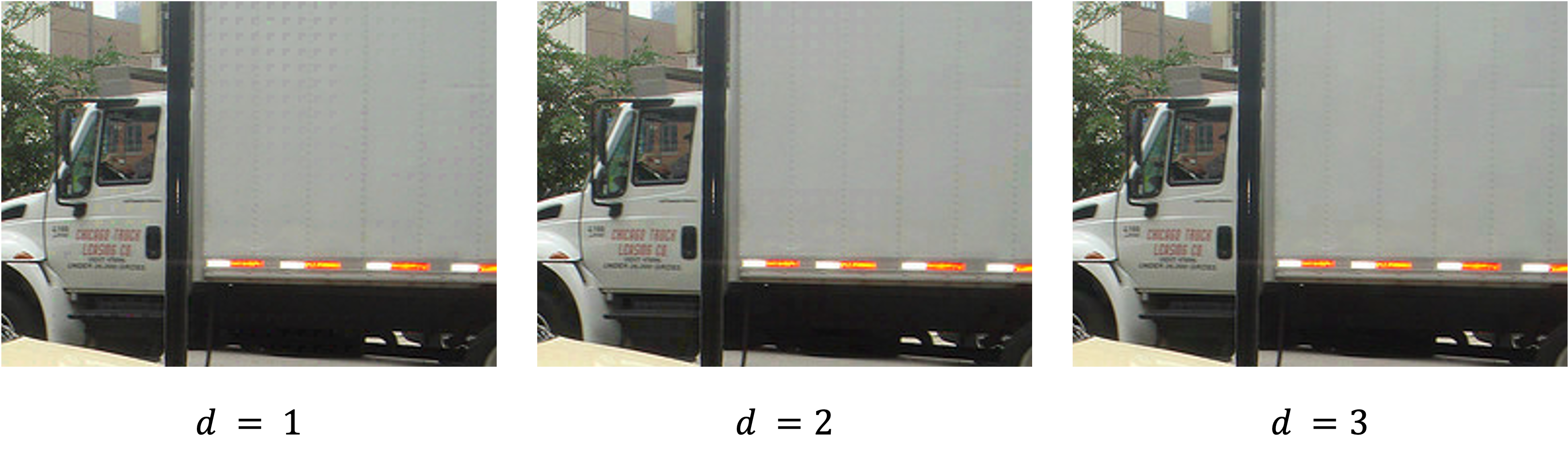} %
    \caption{The same image watermarked with different levels of DWT (using $l=8$). As the number of DWT decomposition levels increases, the embedded watermark becomes more imperceptible to the human eye.}
    \label{fig:diff_dwt_levels} %
\end{figure*}

When employing DWT in image watermarking, it is common practice to apply two or three levels of DWT to enhance robustness \cite{dwt_multilevel, dwt_multilevel_2}. On top of this, we observe that using a higher number of DWT transforms helps to make the watermark more imperceptible, as illustrated in \cref{fig:diff_dwt_levels}, since the embedding is spread more subtly across the transformed coefficients. In \cref{tab:WDR_diff_d}, we report the WDR for different levels of DWT. We observe that the level does not affect the WDR, as all results are at $\approx 100\%$. For this reason, it is generally preferable to choose a larger $d$.
The \textbf{number of levels of DWT} $d$ that can be applied to each sub-block is constrained by the \textbf{sub-block size} $k$. Since each level of DWT decomposition reduces the dimensions of the image (or sub-band) by half in both directions, $k$ must be greater than or equal to $2^d$. Additionally, the \textbf{block size} $m$ must be a multiple of $k$ to ensure proper application of the DWT decomposition to the whole image area.

\begin{table}[h]
\centering
\begin{tabular}{|c|c|c|c|c|}
\hline
\textbf{\# DWT}   & \multicolumn{2}{c|}{\textbf{FPR@1\%}} & \multicolumn{2}{c|}{\textbf{FPR@5\%}} \\
\cline{2-5}
\textbf{levels} & COCO & DALL-E & COCO & DALL-E \\
\hline
$d = 1$ & 100.0\% & 100.0\% & 100.0\% & 100.0\% \\
\hline
$d = 2$ & 100.0\% & 100.0\% & 100.0\% & 100.0\% \\
\hline
$d = 3$ & 99.8\% & 100.0\% & 100.0\% & 100.0\% \\
\hline
\end{tabular}
\caption{We watermark 500 images from both MS-COCO and DALL-E datasets using three different settings for the number of DWT levels $d$. We observe that all settings yield a highly detectable watermark, with detection rates at approximately 100\% for every $d$.} \label{tab:WDR_diff_d}
\end{table}

\subsection{Hypothesis Test for Watermark Detection} \label{app:hyp_test}

To reduce false positives, we use a statistical test per block instead of a fixed 50\% threshold, making the method more robust to natural variation.
Our one-sided hypothesis test  for each block (conducted at a 5\% significance level in our experiments, can be adjusted by the user) is formulated as follows: 
\begin{align*}
H_0:\ &\textit{The block is not watermarked. It violates the red-green} \\
     &\textit{ rule in at least 50\% of its DWT coefficients.} \\
H_1:\ &\textit{The block is watermarked.} 
\end{align*}
The significance level of the hypothesis test can be used to control the false positive rate. Increasing the significance threshold reduces the likelihood of misclassifying unwatermarked blocks as watermarked, which comes at the potential cost of slightly reduced robustness against manipulations.

If the null hypothesis is true, the number of green list coefficients $c_G$ has an expected value of $N/2$ and a variance of $N/4$, where $N$ is the total number of DWT coefficients after $d$ levels of DWT transformations to the block. We can calculate the threshold $c^*_G$ for the number of green list coefficients for which we start rejecting the null hypothesis as:
\begin{align}
c^*_G = \frac{z_{0.95}}{2} \cdot \sqrt{N} + \frac{N}{2}.
\end{align}

\begin{example}
To better understand the impact of scheme parameters on detection, consider example parameter settings. For a block size of $m=96$ and sub-block size of $k=8$ with $d=3$ levels of DWT, this yields a threshold of $c^*_G \approx 82$. This means that for a block to reject the null hypothesis and consider this block watermarked, at least 82 out of 144 DWT coefficients $(>56.94\%)$ need to be from the green domain. 
\end{example}

\subsection{Strengthening against Brute-Force Attacks} \label{private_multiple_keys}

In \cref{sec:embedding} we assume that the user generates a single secret key $\sk$ which we keep fixed for the whole image.
Our system can be strengthened against brute-force attacks by employing $K>1$ distinct secret keys for watermark embedding. 
Instead of using a single secret key for the entire image, the algorithm randomly selects a key from a predefined set for each $m \times m$ block.  
During detection, we evaluate the image using all $K$ possible keys and correct for multiple hypotheses via Bonferroni correction. %
When partitioning DWT coefficients, half of the intervals are defined as allowed regions, meaning that each coefficient will be in the allowed domain for $K/2$ keys and in the disallowed domain for the other $K/2$ \cite{kirchenbauer2024watermarklargelanguagemodels}. This approach helps prevent statistical attacks since no consistent patterns can be easily discovered when analysing large numbers of watermarked images, even when an attacker attempts to aggregate statistics across multiple images.

\subsection{Adaptive Embedding Based on Block Entropy} \label{adapt-block-entropy}

The interval length $l$ of our watermark presents a trade-off between visibility and robustness. Higher values of $l$ result in watermarks that are more robust to image manipulations but may become visible in low-entropy regions (see \cref{fig:diff_interval_lengths}). To preserve this robustness while maintaining imperceptibility, we propose to embed the watermark only in high-entropy areas. 
For the experiments in \cref{sec:robustness} we implement the following procedure: Before executing line 5 of \cref{alg:MELB} (and for the detection before line 12 of \cref{alg:MELB_detection}), we calculate the entropy $H(b)$ of each block $b$ according to \cref{entropy_eq}. Subsequently, we determine the entropy threshold $\tau = \text{median}\{H(b_1), \dots, H(b_n)\}$ where $n$ is the total number of blocks in the image.  Only blocks with entropy above the threshold ($H(b) > \tau$) are processed for watermark embedding or detection. Note that while we use the median as our threshold, alternative percentile thresholds could be selected by the user depending on the desired embedding capacity and imperceptibility. Excluding low-entropy regions from watermarking is generally reasonable, as these areas typically correspond to monochromatic or background regions with minimal information rather than the actual content we seek to protect. This approach aligns with similar entropy-based filtering strategies employed in previous work \cite{kirchenbauer2024watermarklargelanguagemodels}. 

\subsection{Crop-Resilient Detection} \label{sec:cropping}

\begin{figure}
    \centering
    \includegraphics[width=0.9\columnwidth]{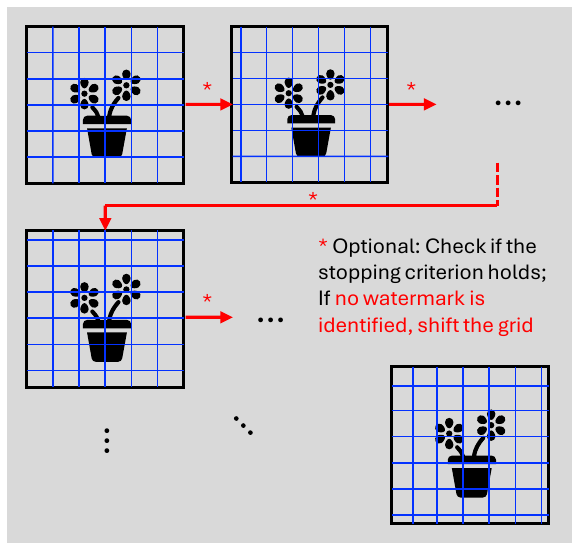} %
    \caption{To make our watermark detection method resilient to cropping, we implement a brute-force approach to find the original position of the grid. An early stopping criterion can be used for computational efficiency. }
    \label{fig:cropping_idea} %
\end{figure}

Since we embed the watermark into blocks throughout the image, we can implement a detection method that identifies watermarks even in cropped images. As visualised in \cref{fig:cropping_idea}, this detection is based on a brute-force strategy to systematically search for the correct starting point of the embedded blocks. 

To detect a watermark in a potentially cropped image, we iterate through the image in $m\times m$ pixel blocks. Since the original starting point of the watermark grid is unknown in a cropped image, we exhaustively test all possible starting positions for the first $m \times m$ block. For the detection process, we propose two different strategies:
\begin{enumerate}
    \item \textbf{Score-based selection:}  We collect the detection scores for all possible grid alignments and output the highest identified score. 
    \item \textbf{Early stopping via fixed threshold:} We choose a detection threshold calibrated to our desired FPR and stop the search if this threshold is met. 
    \item \textbf{Early stopping via hypothesis testing:} We evaluate the percentage of watermarked blocks for each iteration, performing a one-sided hypothesis test to determine if the proportion of watermarked blocks exceeds 50\% at a specified significance level. If this threshold is met, we conclude that a watermark has been detected and stop the search.  
\end{enumerate}
This approach requires a maximum of $m^2$ iterations to cover all possible starting positions, which is computationally practical for the empirically effective block sizes we consider ($m = 96$).

\begin{figure}
    \centering
    \includegraphics[width=\columnwidth,clip,trim=0 2cm 0 0]{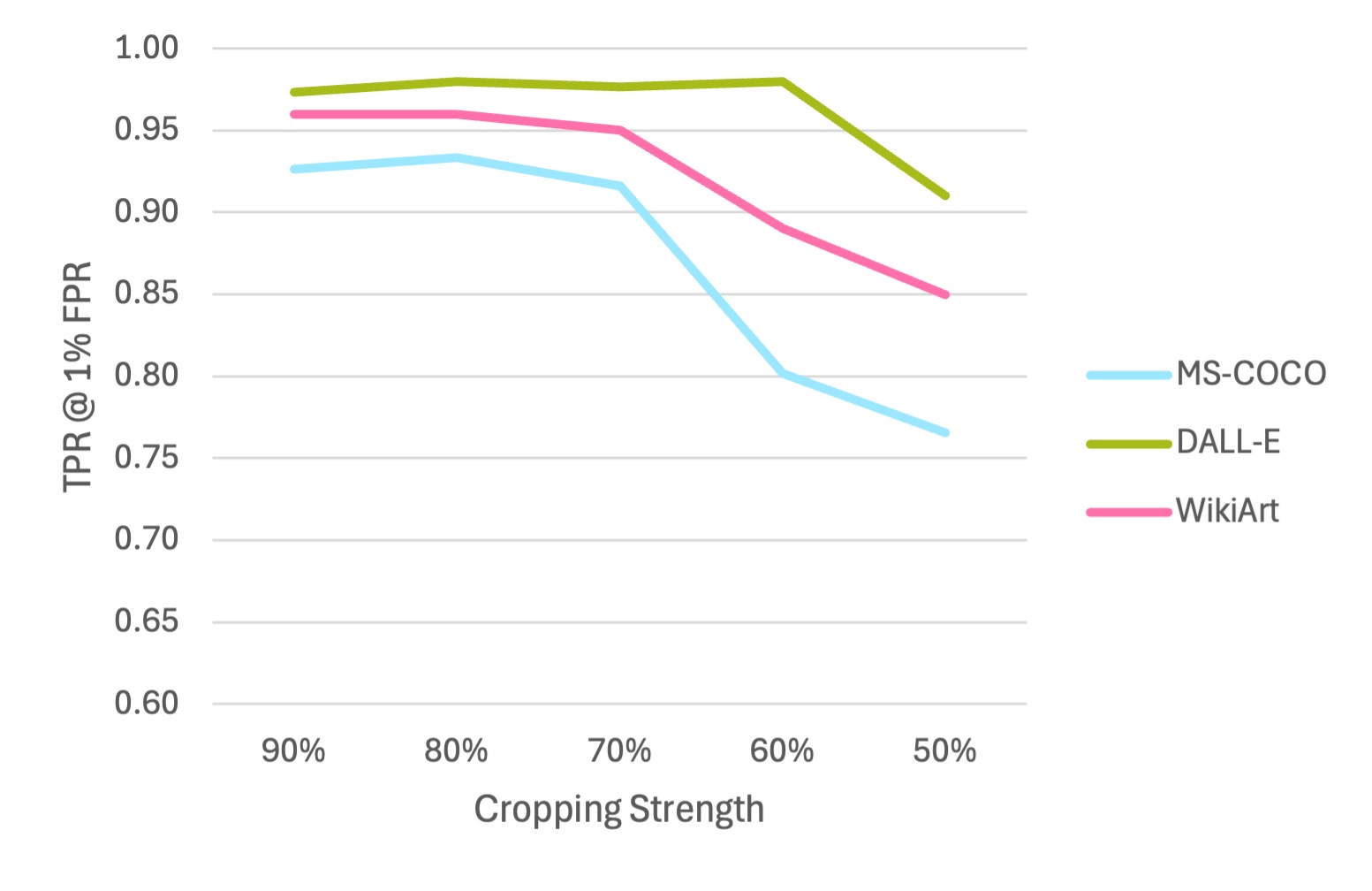} %
    \caption{We evaluate the TPR at 1\% FPR under varying cropping strengths across datasets. Despite increasing image loss, the watermark remains detectable, with performance remaining strong even when up to 50\% of the image is cropped.}
    \label{fig:cropping_results} %
\end{figure}

\section{Further Experimental Details}

\subsection{Embedding Settings} \label{app:embedding_settings}

We apply $d=3$ levels of DWT, and accordingly set the block size to the smallest allowed value $k = 2^d = 8$. We partition the domain of the DWT coefficients into intervals of size $l$ over the range $[-3000, 3000]$ and treat values outside of this range as allowed (green). To minimise the visible impact of the introduced perturbations after reversing the transform, we restrict the maximum perturbation to at most $\pm 3l$. If no green intervals are found within this range, the coefficient remains unchanged. This restriction is justified as the probability of all six intervals (left and right for each of the 3~channels) surrounding a coefficient being red is very low, at approximately $\frac{1}{2}^6 \approx 0.016$. 

For the detectability, image forensics, and cropping experiments (Sections \ref{sec:exp_detect}, \ref{sec:im_forensics}, and \ref{sec:exp_cropping}), we set the interval length to $l=8$. To assess the method's behaviour under geometric transformations (\cref{sec:geom_attacks}), we increase the interval length to $l=14$ and selectively embed the watermark in high-entropy regions. Specifically, we calculate the median entropy across all blocks and embed the watermark only in blocks with entropy values above this median threshold, thereby excluding regions with very low entropy values. 

\subsection{Choosing the Detection Threshold}

For a direct comparison with the results presented in \cite{zodiac}, we follow their methodology: For watermarking techniques that embed bitstrings, detection thresholds are defined to reject the null hypothesis (\ie, the absence of a watermark) at $p<0.01$. This requires the correct detection of at least 24 out of 32 bits for 32-bit methods, and 61 out of 96 bits for StegaStamp.
For score-based watermarking methods, the threshold parameter $p$ is calibrated to achieve reasonable FPRs. Following this approach, our score-based method is calibrated to match similar FPR values as in \cite{zodiac} across different datasets. Through empirical evaluation of various threshold settings (\cref{tab:thresholds}), optimal threshold values were determined: $p = 0.37$ for the COCO dataset, $p = 0.36$ for Diffusion DB, and $p = 0.46$ for WikiArt.

\begin{table}[h!]
  \centering
  \resizebox{\columnwidth}{!}{%
    \begin{tabular}{|ccc|ccc|ccc|}
      \hline
      \multicolumn{3}{|c|}{\textbf{MS-COCO}} 
        & \multicolumn{3}{|c|}{\textbf{DiffusionDB}} 
        & \multicolumn{3}{c|}{\textbf{WikiArt}} \\
      \hline
      $\tau$ & FPR   & WDR   & $\tau$ & FPR   & WDR   & $\tau$ & FPR   & WDR   \\
      \hline
      \cellcolor{gray!30}\textbf{0.37} & \cellcolor{gray!30}\textbf{0.062} & \cellcolor{gray!30}\textbf{1.000}
        & 0.35 & 0.052 & 1.000
        & 0.45 & 0.067 & 1.000 \\
      0.40 & 0.046 & 1.000
        & \cellcolor{gray!30}\textbf{0.36} & \cellcolor{gray!30}\textbf{0.047} & \cellcolor{gray!30}\textbf{1.000}
        & \cellcolor{gray!30}\textbf{0.46} & \cellcolor{gray!30}\textbf{0.060} & \cellcolor{gray!30}\textbf{1.000} \\
      0.50 & 0.014 & 1.000
        & 0.40 & 0.028 & 1.000
        & 0.50 & 0.039 & 1.000 \\
      0.60 & 0.005 & 0.998
        & 0.50 & 0.009 & 1.000
        & 0.65 & 0.006 & 1.000 \\
      \hline
    \end{tabular}%
  }
  \caption{FPR and WDR for MS-COCO, DiffusionDB, and WikiArt at different detection thresholds $\tau$.}
  \label{tab:thresholds}
\end{table}

\section{Additional Figures and Analysis}

In this section, we provide more details on our evaluations and additional figures.

\subsection{Image Quality of Manipulated Images} \label{app:attack-image-quality}

\begin{figure*}[h]
    \centering
    \includegraphics[width=\textwidth, height = 11cm]{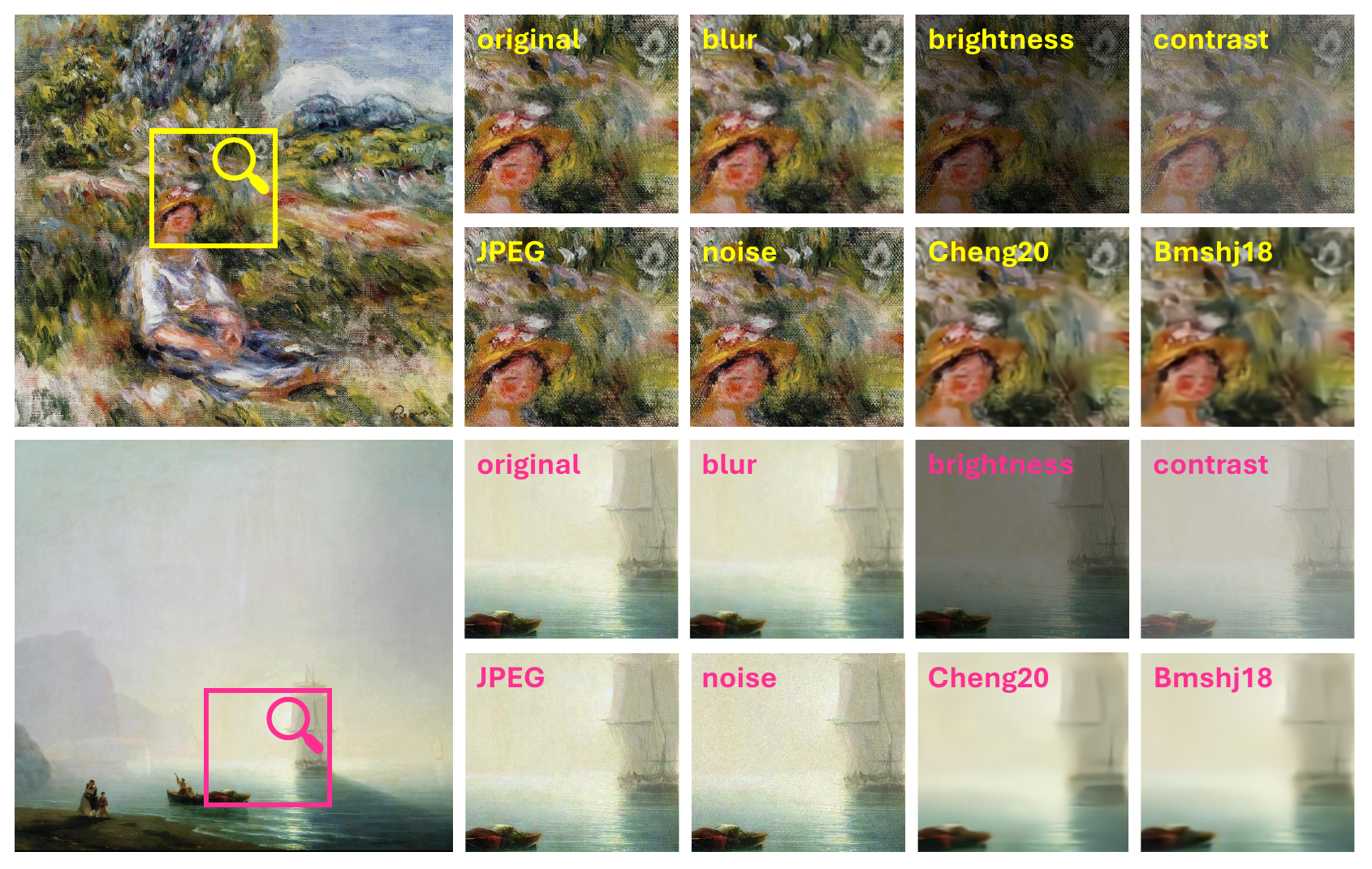}
    \caption{The effectiveness and visibility of attacks can vary across different images. We illustrate two examples - one highly resilient to visual changes, and one more sensitive. Each image is subjected to the full set of modifications described in \cref{sec:attack_methods}. This figure compares the impact of different attacks on these images. The upper image contains rich visual detail and texture, which benefits the malicious actor by making perturbations less noticeable to the human eye. In contrast, the lower image is relatively smooth and monochromatic, offering fewer opportunities to conceal changes. As a result, even small amounts of added noise become clearly visible in the lower image, whereas they remain imperceptible in the upper one. Nonetheless, for both images, we observe that certain transformations, such as brightness, contrast, and the attacks Cheng20 and Bmjsh18, significantly degrade visual quality.}
    \label{fig:attacks_vis} %
\end{figure*}

We first illustrate two example images to demonstrate how the same set of modifications can have different visual effects depending on the input image. One of the images is more robust to the changes, while the other reacts more sensitively, as shown in \cref{fig:attacks_vis}.

\begin{figure*}
    \centering
    \includegraphics[width=\textwidth]{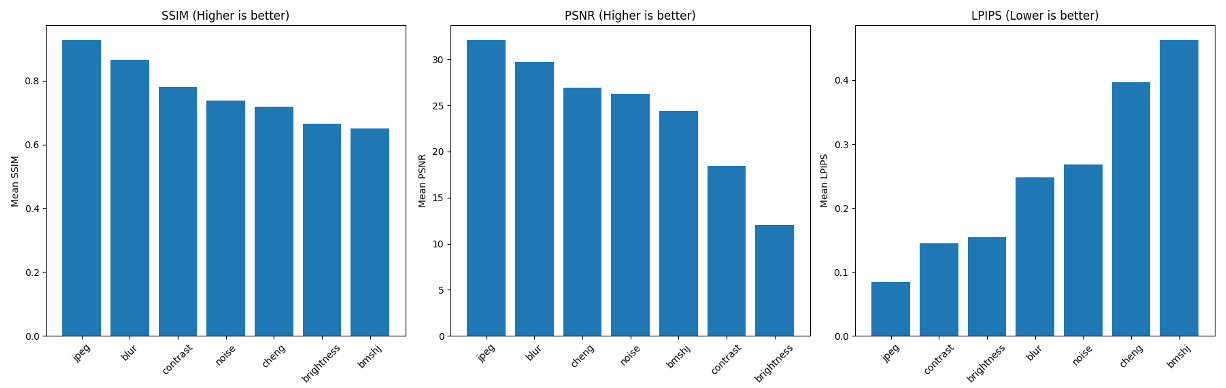} %
    \caption{Similarity scores (SSIM, LPIPS, PSNR) for each attack, averaged over all datasets. Since these metrics capture different aspects of image similarity, we obtain varying rankings across attacks. Still, we can observe clear patterns, such as JPEG being consistently ranked first, meaning that it is causing the least degradation, while Bmjsh18 ranks among the most damaging.}
    \label{fig:metrics_sep} %
\end{figure*}

Based on this observation, we further quantify the variability in modification effects by evaluating each manipulation using a set of similarity metrics. \cref{fig:metrics_sep} presents the results for each similarity metric individually. It is evident that some manipulations, such as JPEG compression, are consistently less perceptible across all metrics, while others (\eg, Bmjsh18) cause significant disruptions to image similarity. As each metric captures a different aspect of perceptual similarity, we cannot observe a consistent ranking of manipulations based on image quality degradation. 

\begin{figure}[t]
    \centering
    \includegraphics[width=\columnwidth]{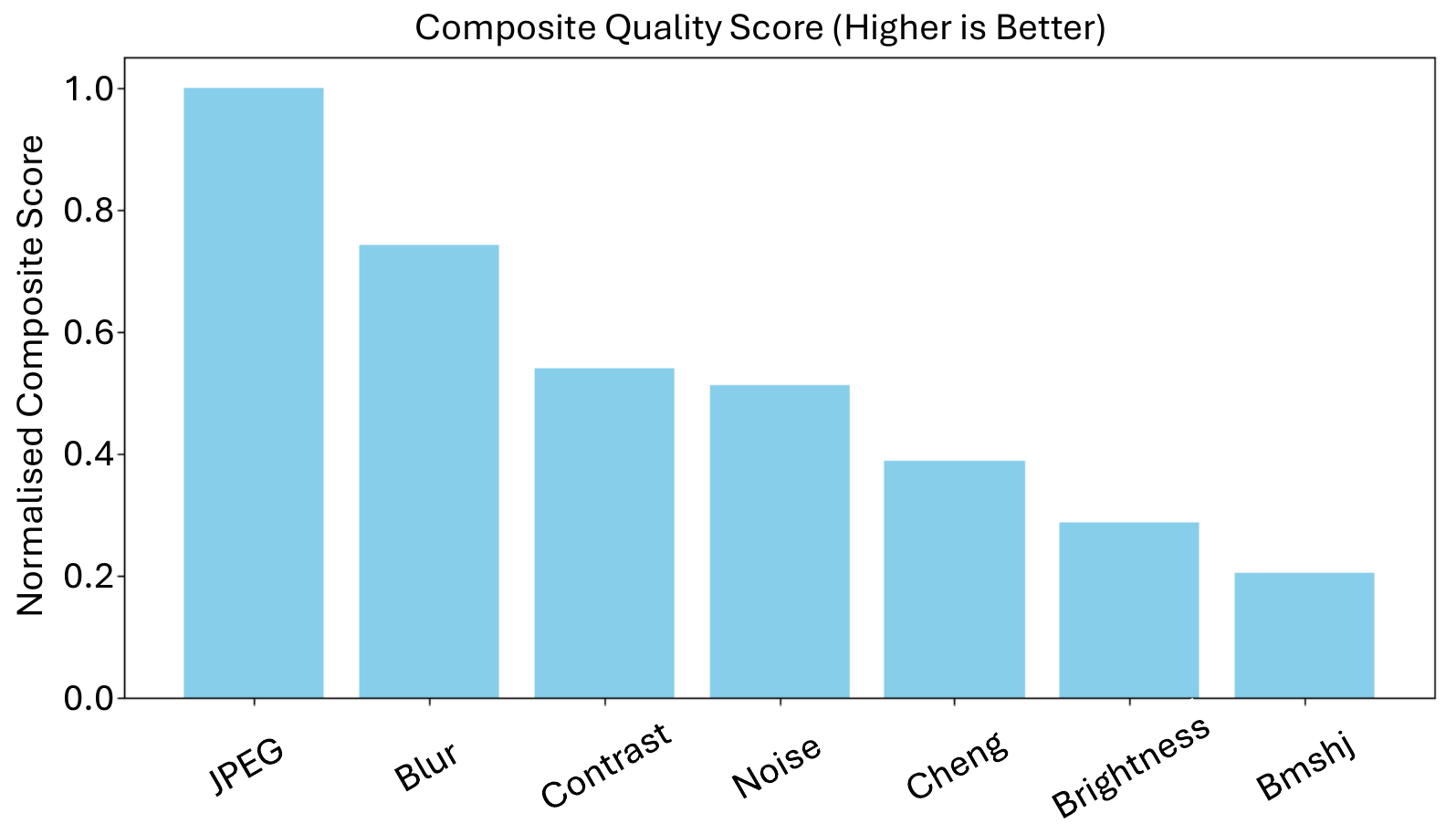} %
    \caption{Combined similarity score for each attack, aggregating normalised SSIM, LPIPS, and PSNR scores.  Higher scores indicate greater similarity to the original image and therefore less visible quality degradation.}
    \label{fig:metrics_combined} %
\end{figure}

To provide a more comprehensive overview, we normalise all metric scores to the range $[0, 1]$ and combine them into a unified scale, shown in \Cref{fig:metrics_combined}. As the transformations Cheng20, Bmshj18 and Brightness failed to achieve 50\% of the average quality metric, we deemed these to be insufficient to achieve the attacker's goals and so did not consider them further.

\subsection{Distribution of Image Sizes} \label{app:image_sizes}

\begin{figure}[t]
    \centering
    \includegraphics[width=\columnwidth]{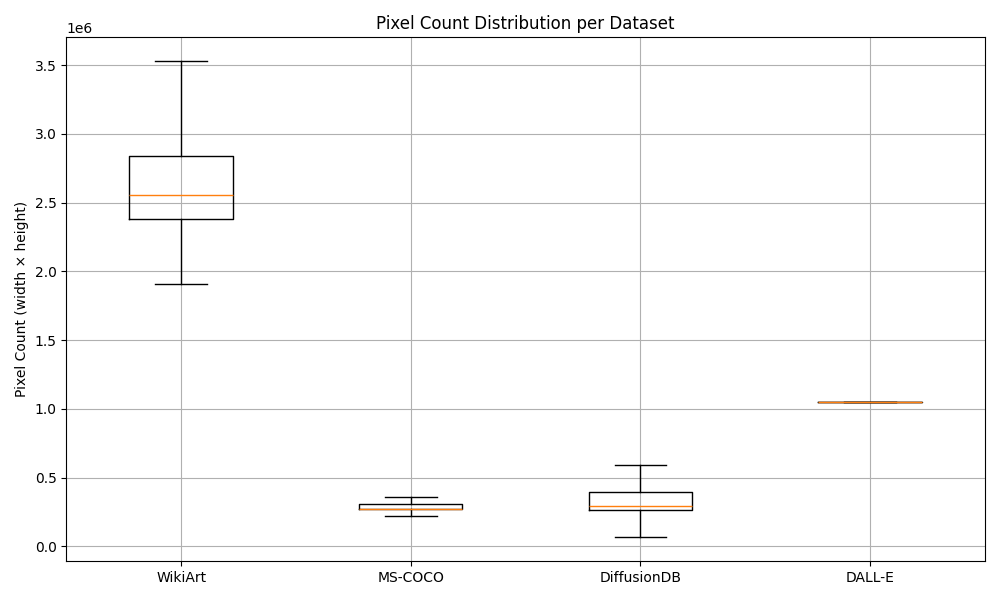} %
    \caption{Pixel count distribution across all evaluation datasets used in our experiments.}
    \label{fig:pixel_boxplot} %
\end{figure}

To ensure diversity in our evaluation, we include datasets with varying image sizes. WikiArt images are the largest, with a median size over $9\times$ larger than MS-COCO and nearly $2.5\times$ larger than DALL-E. DALL-E and DiffusionDB represent mid-sized images, with DALL-E being notably consistent in resolution, while DiffusionDB shows broader variability. An overview of the pixel count distributions is shown in Figure~\ref{fig:pixel_boxplot}.

When it comes to image forensics and analysing the detection map, our watermarking method performs best when applied to large images. In these cases, it becomes easier to analyse the patterns of red squares in the localised detection map. The larger scale helps us distinguish between actual modifications and random false positives, which can occur in small numbers even in unaltered images.

\subsection{Additional Robustness Evaluations}

In this section, we provide additional evaluation of our watermarking method's robustness against various image manipulations. To complement the cropping analysis presented in \cref{sec:exp_cropping}, we include \cref{fig:cropping_results} for completeness, which visualises the robustness of our method to different levels of cropping.
In addition to the results presented in \cref{sec:geom_attacks}, we assess the detection performance of our method under various image modifications:

\begin{itemize}
    \item JPEG compression with quality factors 30, 50, 70 and 90.
    \item Gaussian blur with standard deviation 1 and kernel sizes of 3, 5 and 7.
    \item Adding Gaussian noise with standard deviation $\sigma$ $\in \{ 0.02,$ $ 0.04, 0.05, 0.06, 0.08\}$.
    \item Contrast and brightness adjustment with factors $\in \{0.5, 0.8,$ $ 1.0, 1.2, 1.6, 2.0\}$.
\end{itemize}

\begin{figure*}
    \centering
    \begin{subfigure}[b]{0.32\textwidth}
        \includegraphics[width=\textwidth, trim={1.2cm 0.5cm 1.2cm 1.2cm}, clip]{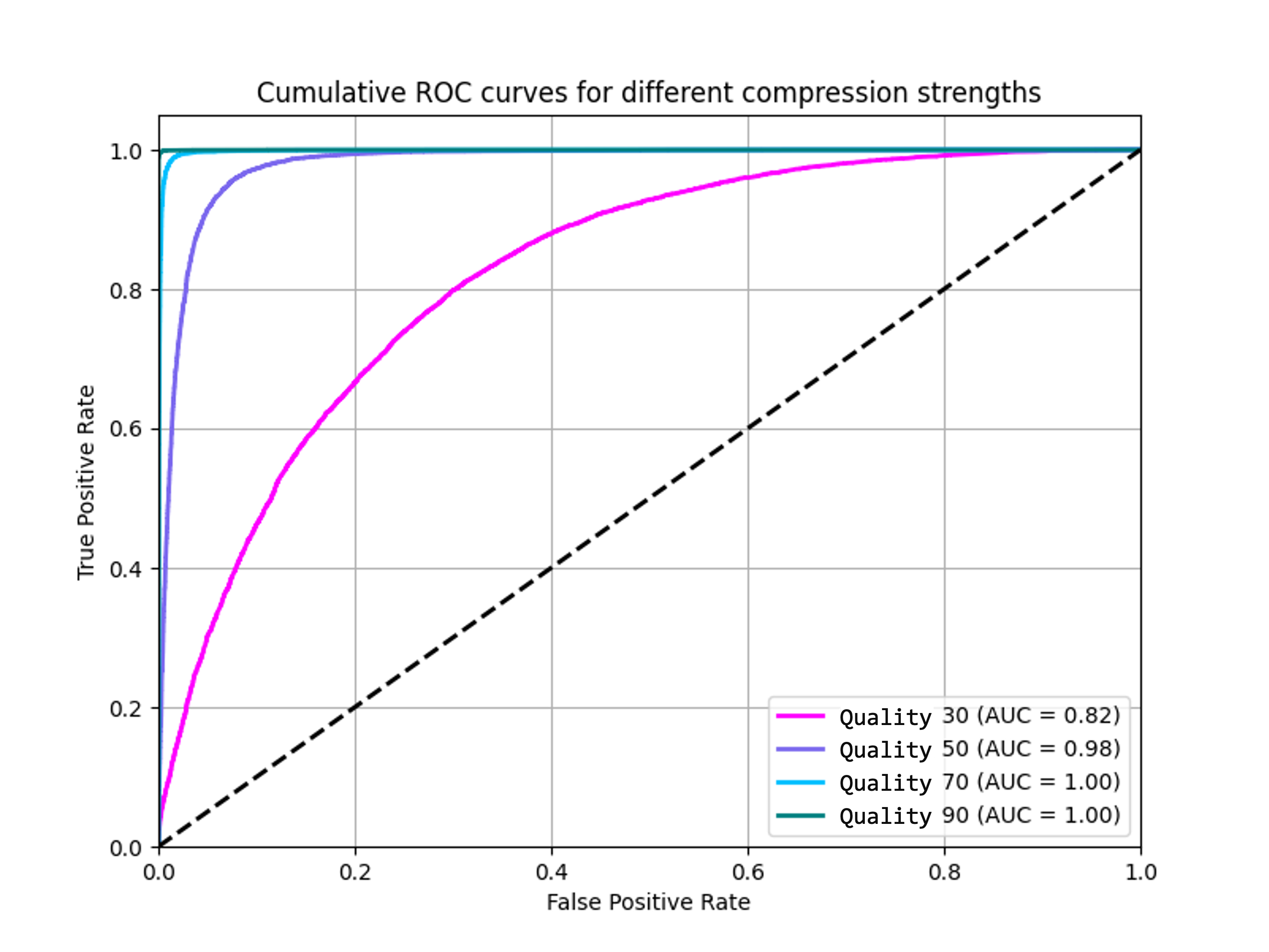}
        \caption{JPEG compression}
        \label{fig:roc_jpeg}
    \end{subfigure}
    \hfill
    \begin{subfigure}[b]{0.32\textwidth}
        \includegraphics[width=\textwidth, trim={1.2cm 0.5cm 1.2cm 1.2cm}, clip]{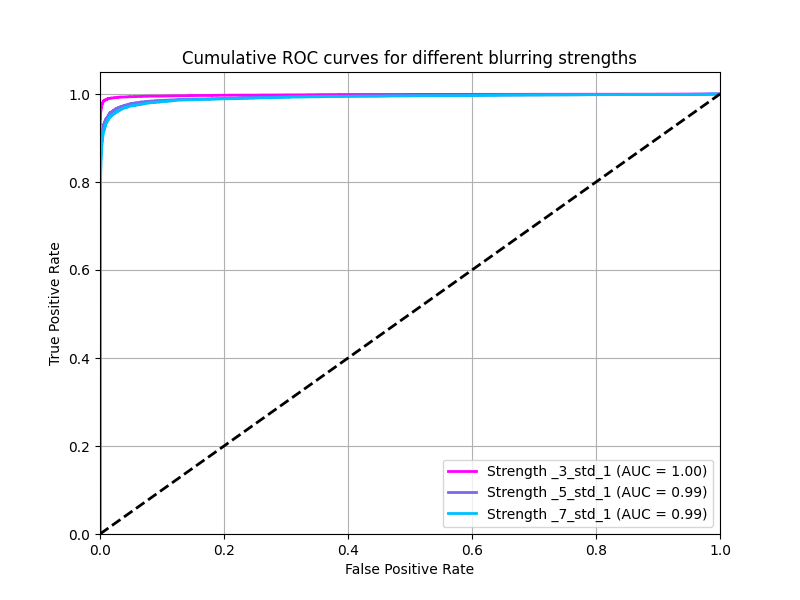}
        \caption{Blurring}
        \label{fig:roc_blur}
    \end{subfigure}
    \hfill
    \begin{subfigure}[b]{0.32\textwidth}
        \includegraphics[width=\textwidth, trim={1.2cm 0.5cm 1.2cm 1.2cm}, clip]{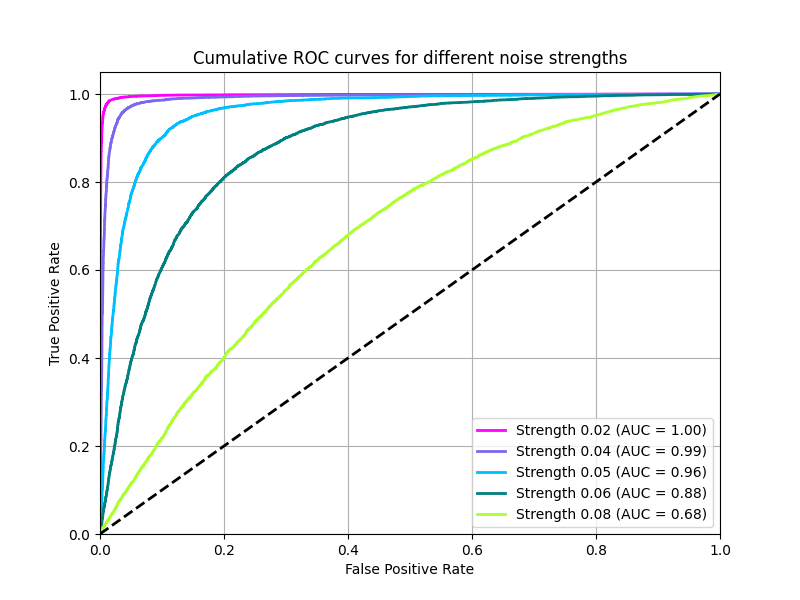}
        \caption{Noise}
        \label{fig:roc_noise}
    \end{subfigure}

    \vspace{0.5cm} %

    \begin{subfigure}[b]{0.32\textwidth}
        \includegraphics[width=\textwidth, trim={1.2cm 0.5cm 1.2cm 1.2cm}, clip]{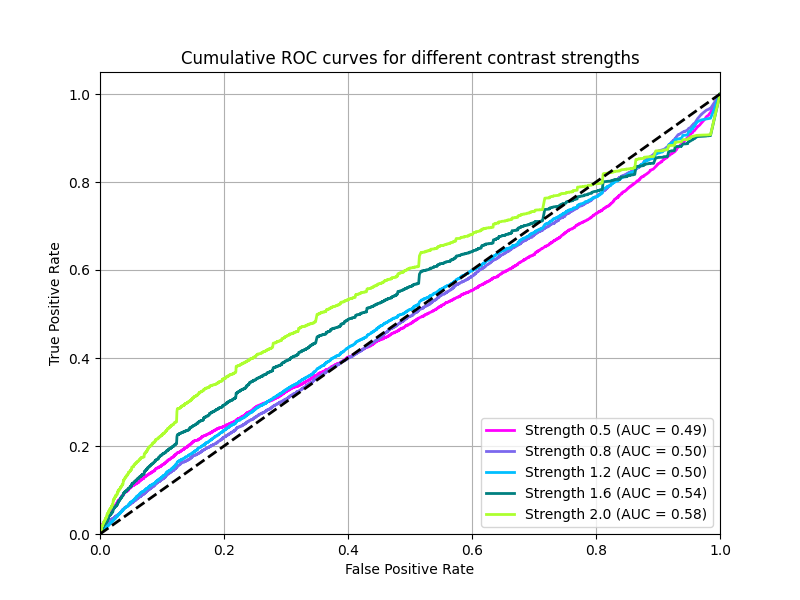}
        \caption{Contrast}
        \label{fig:roc_contrast}
    \end{subfigure}
    \hfill
    \begin{subfigure}[b]{0.32\textwidth}
        \includegraphics[width=\textwidth, trim={1.2cm 0.5cm 1.2cm 1.2cm}, clip]{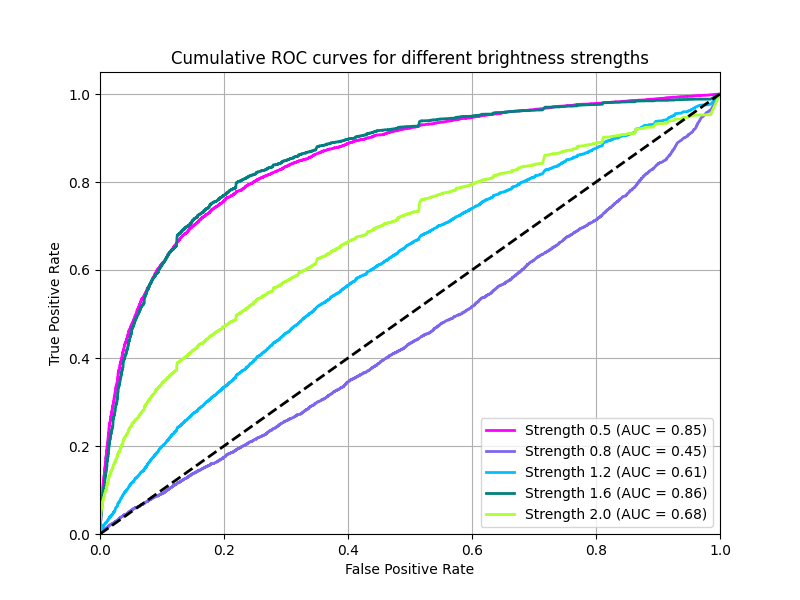}
        \caption{Brightness}
        \label{fig:roc_brightness}
    \end{subfigure}
    \hfill
    \begin{subfigure}[b]{0.32\textwidth}
        \centering
        \textcolor{white}{empty} %
    \end{subfigure}
    \caption{Cumulative ROC curves for MS-COCO, DiffusionDB and WikiArt.}
    \label{fig:COCO_roc_curves}
\end{figure*}

We present ROC curves for each type of distortion in \cref{fig:COCO_roc_curves}. We can observe that our watermarking scheme exhibits high robustness against JPEG compression. Detection performance is nearly perfect for quality factors 90, 70, and 50, with AUC values close to 1.00. Even under strong compression with a quality factor of 30, the method maintains good detectability, achieving an AUC of 0.83.

Similarly, the method shows excellent robustness to Gaussian blurring, as depicted in \cref{fig:roc_blur}. For a kernel size of 3, the AUC is 1.00, indicating perfect detection. Increasing the kernel size to results in only a marginal decrease in performance (AUC of 0.99).

The performance under additive Gaussian noise is evaluated in \cref{fig:roc_noise}. We can observe that the watermark detection remains high as the noise strength increases further to $\sigma = 0.06$ (AUC = 0.91) and $\sigma = 0.08$ (AUC = 0.81).

Contrast and brightness adjustments, however, pose a greater challenge (see \cref{fig:roc_contrast} and \cref{fig:roc_brightness}. The detection performance is weaker across all tested factors. These image modifications directly modify the pixel intensity values and the overall dynamic range of the image. Given that the DWT coefficients are computed from these pixel values, substantial alterations to the input data necessarily lead to significant changes in the calculated coefficient magnitudes and distributions.

Overall, this additional evaluation indicates that our watermarking method is highly robust to common distortions like JPEG compression, Gaussian blur, and additive Gaussian noise. While detection performance can be impacted by strong contrast and brightness adjustments, it is noteworthy that such significant modifications typically result in substantial and perceptible visual degradation of the image content. Consequently, the practical relevance of this sensitivity might be reduced in scenarios where maintaining the visual authenticity and quality of the content is important.

\subsubsection{Attacking the Seed} \label{sec:attack_seed}
To evaluate the robustness of the detection process of our watermarking scheme, we design a straightforward attack targeting the sensitivity of the block seed ($\getSeed$ function in \cref{alg:MELB,alg:MELB_detection}, see \cref{alg:get_seed} for a definition). The attack aims to modify the mean colour of each block, which is crucial for watermark detection. Specifically, we explore three variations of this attack:
1) Adjusting the mean colour to the \textbf{nearest value} that would round to a different integer when using our rounding-to-30 scheme.
2) Increasing the mean colour to the \textbf{next higher value} that would result in a different rounded integer.
3) Decreasing the mean colour to the \textbf{next lower value} that would yield a different rounded integer.

\begin{figure*}
    \centering
    \begin{subfigure}[b]{0.45\textwidth}
        \includegraphics[width=\textwidth, height = 7.6cm]{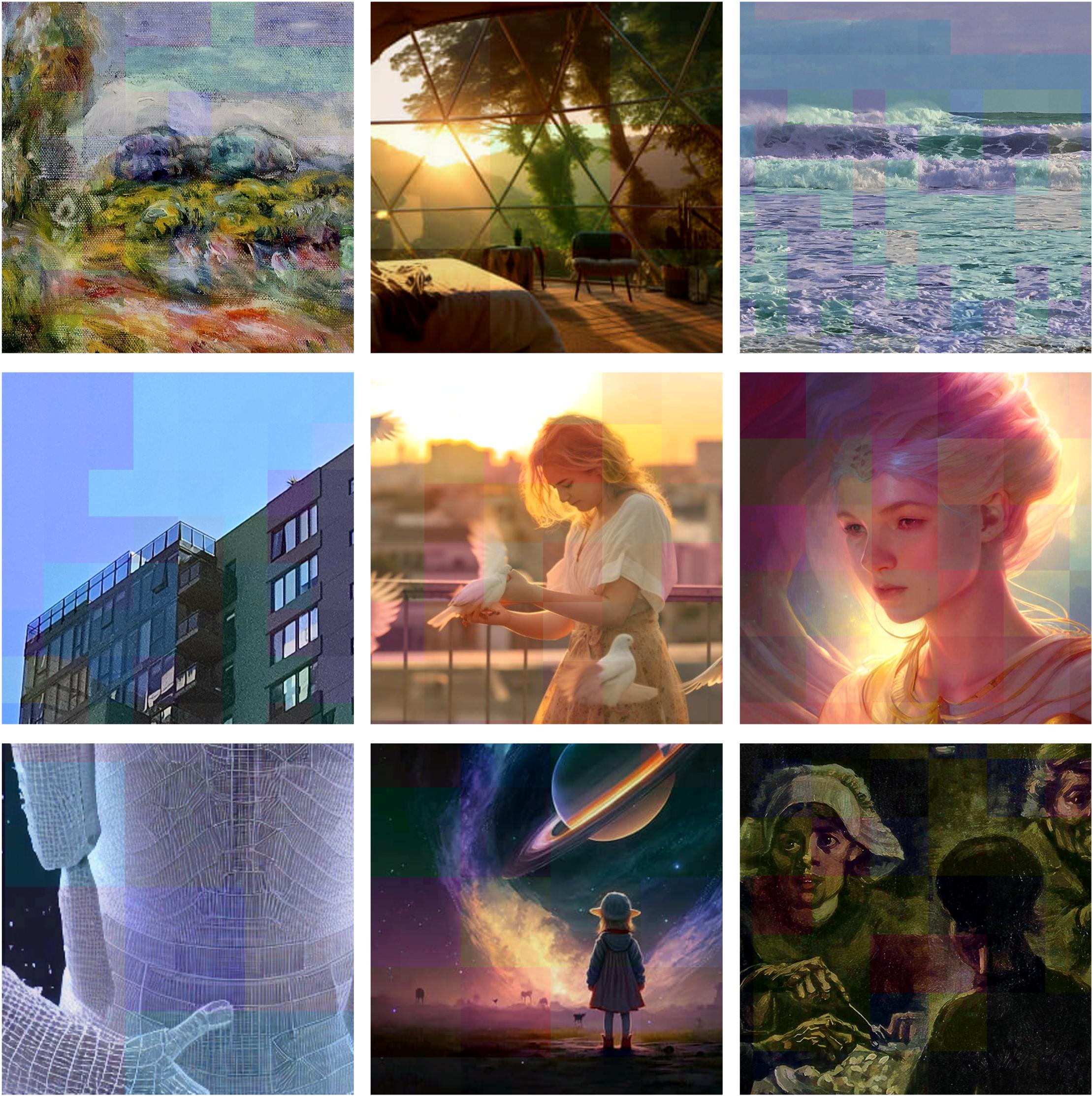}
        \caption{Attacked images}
        \label{fig:seed_att}
    \end{subfigure}
    \hfill
    \begin{subfigure}[b]{0.45\textwidth}
        \includegraphics[width=\textwidth, height = 7.6cm]{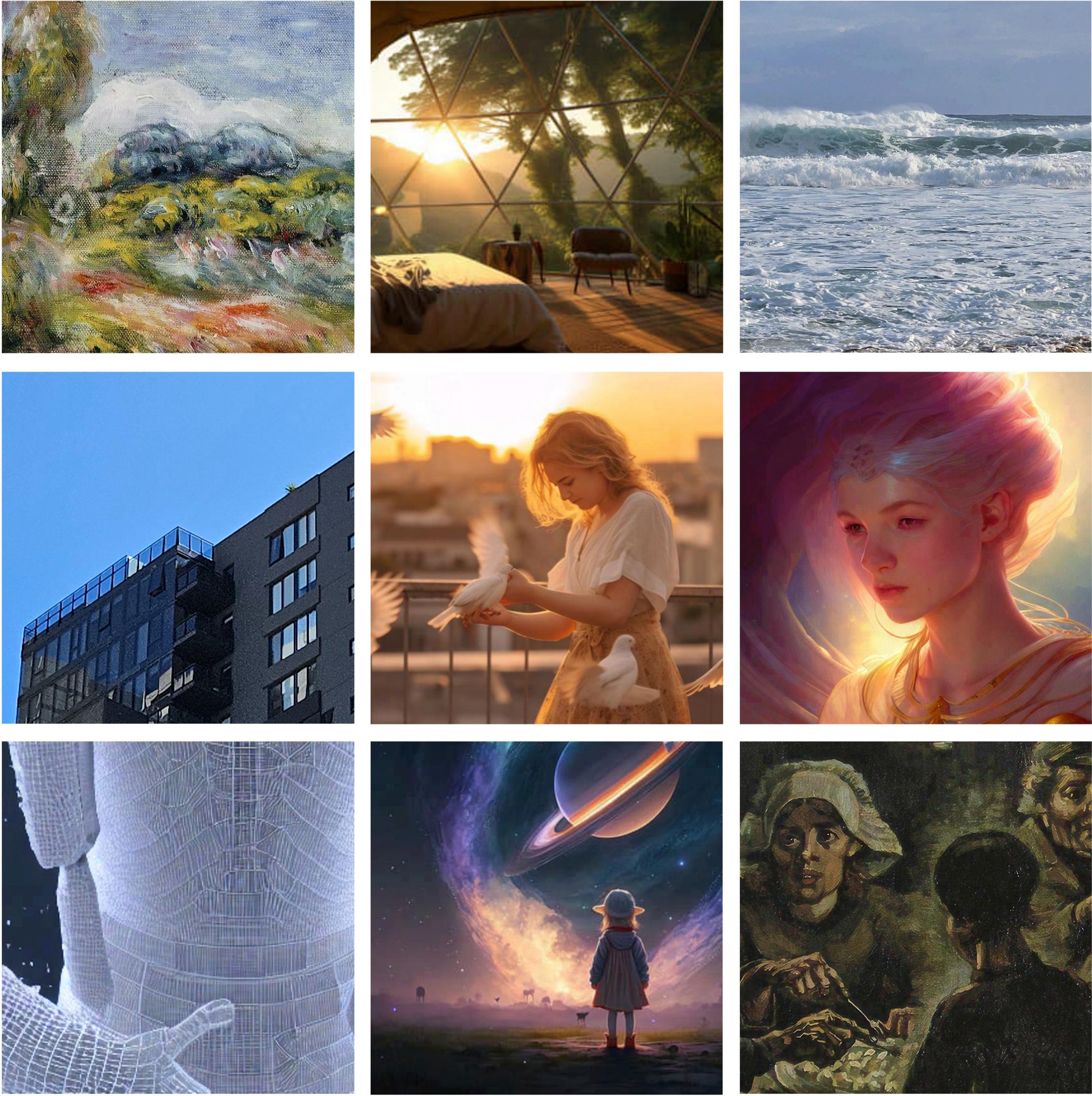}
        \caption{Original images}
        \label{fig:seed_orig}
    \end{subfigure}

    \caption{Evaluation of seed robustness against mean colour alteration attacks (\cref{sec:attack_seed}). \cref{fig:seed_orig} shows the watermarked image, where the partitioning seed during embedding relies on the block mean colour rounded with $r=30$. \cref{fig:seed_att} shows the watermarked image after the attack aimed at modifying block mean colours to disrupt seed reconstruction. The visible degradation in \cref{fig:seed_att} demonstrates that altering the seed sufficiently to hinder watermark detection requires introducing visually significant artifacts.}
    \label{fig:combined_images}
\end{figure*}

In \cref{fig:combined_images} we demonstrate example outputs from this attack, indicating that all three variations of this attack necessitated substantial alterations to the mean colour values. These changes resulted in visual artefacts, that significantly compromise image quality. 
These results demonstrate the difficulty of watermark removal by manipulating the block seeds. 

\section{Societal Impact}

Watermarking improves the traceability and authenticity of digital content, including AI-generated media. This capability offers important societal benefits, such as protecting intellectual property and helping detect misinformation or manipulated content. This traceability can also have unintended consequences, including privacy concerns and potential misuse, such as unauthorised watermarking of unwatermarked content to falsely claim ownership. Our approach mitigates forgery risks through secret-key-based pseudorandom embedding (see \cref{remark 1}), and the localised detection map provides transparency to aid forensic verification while recognising limitations such as possible false detection errors (\cref{sec:Map}). Overall, watermarking plays a crucial role in supporting digital content authenticity and traceability, making it a valuable tool. We believe the societal benefits of watermarking in maintaining trust and authenticity outweigh potential risks.

\end{document}